\documentclass[10pt,twocolumn,letterpaper]{article}

\usepackage[final]{cvpr}      
\definecolor{cvprblue}{rgb}{0.21,0.49,0.74}
\usepackage[pagebackref,breaklinks,colorlinks,allcolors=cvprblue]{hyperref}
\usepackage{pifont}
\usepackage{multirow}
\usepackage[table,x11names]{xcolor}
\usepackage{amsthm}
\newtheorem{theorem}{Theorem} 
\newtheorem{definition}{Definition}
\newtheorem{lemma}{Lemma}[theorem]

\def\paperID{19097} 
\def\confName{CVPR}
\def\confYear{2026}

\title{MemFlow: A Lightweight Forward Memorizing Framework for Quick Domain Adaptive Feature Mapping}

\author{
Jianming Lv$^{1}$\thanks{Corresponding author :jmlv@scut.edu.cn} \and 
Chengjun Wang$^{1}$ \and 
Depin Liang$^{1}$ \and 
Qianli Ma$^{1}$ \and
Wei Chen$^{2}$ \and
Xueqi Cheng$^{2}$
\\
$^{1}$South China University of Technology \\
$^{2}$Institute of Computing Technology, Chinese Academy of Sciences \\
{\tt\small \{jmlv, qianlima\}@scut.edu.cn \tt\small \{cswangchengjun, csldp\}@mail.scut.edu.cn} \\
\tt\small \{chenwei2022, cxq\}@ict.ac.cn}

\begin{document}
\maketitle
\vspace{-0.5cm}
\begin{abstract}
	Deploying pretrained visual models in real-world environments often suffers from significant performance degradation due to the diversity of testing scenarios. Continuous adaptation of learning models on edge devices via unlabeled data collected from the target domain is highly effective for boosting generalization capability. However, gradient-backpropagation-based optimization of the massive parameters in deep neural networks is vastly more time-consuming than forward inference, rendering online learning infeasible on low-power edge devices. To address this critical challenge, we propose a lightweight gradient-free forward-memorizing framework, namely MemFlow, which leverages a frozen backbone and enables efficient fine-tuning of the mapping between features and predictions. Specifically, MemFlow employs randomly connected neurons to memorize feature-label associations; within the network, spiking signals are propagated, and predictions are generated by associating neuron-stored memories according to their confidence levels. More notably, MemFlow supports reinforced memorization of feature mappings using unlabeled data, thereby enabling rapid adaptation to new domains. Extensive experiments on four real-world cross-domain datasets demonstrate that MemFlow achieves performance improvements of up to 10\% while consuming less than 1\% of the computational time required by traditional domain adaptation methods. \href{https://github.com/so-link/MemFlow}{The code is available at https://github.com/so-link/MemFlow}
\end{abstract}

\begin{figure}[tb]
	\vskip 0.2in
	\centering
    \vspace{-1cm}
    \hspace{-0.5cm}
	\includegraphics[width=1.05\linewidth]{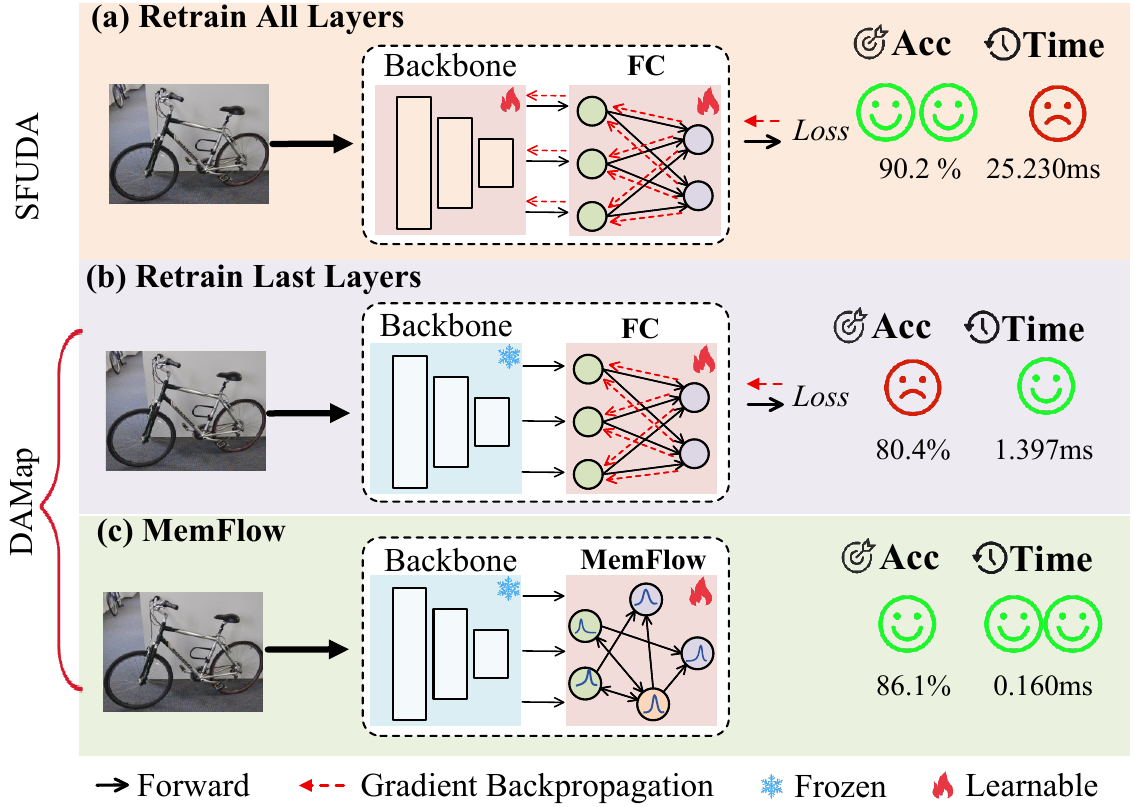}
	\caption{Illustration about the difference between  SFUDA and DAMap methods. The testing accuracy and adaptation time per instance  are tested in  Office-31 dataset.}  
    \vspace{-0.5cm}
    \label{fig.1}
\end{figure}

\section{Introduction}
\label{Intro}
Most deep Artificial Neural Networks (ANNs) rely on large-scale labeled data to achieve superior performance. They also tend to overfit training datasets during the optimization of deep networks with numerous parameters. Particularly in visual classification tasks, direct cross-domain transfer of deep models often results in substantial degradation in performance \cite{donahue2014decaf, gulrajani2020search}.

Recently, numerous Source-free Unsupervised Domain Adaptation (SFUDA) methods have been proposed to utilize unlabeled data from the target domain for continuous model optimization. Specifically, pseudo-label-based methods \cite{liang2021domain,litrico2023guiding, wang2025vicinal}  assign pseudo-labels to target domain data, enabling model optimization in a supervised manner. Clustering-based methods \cite{liang2020we,li2021cross, pan2025overcoming} leverage relationships between unlabeled samples to fine-tune deep models. 


However, gradient-backpropagation-based optimization of deep neural networks in these SFUDA methods is typically much more time-intensive than the inference process (Fig.~\ref{fig.1}a). This undermines the feasibility of continuous learning on resource-constrained edge devices. 

A critical question thus arises:  can we freeze the deep backbone responsible for feature extraction and only fine-tune the feature-prediction mapping to efficiently boost performance on the unlabeled target domain? This constitutes a notable and practical challenge, which we formally define as the \textbf{Domain Adaptive Feature Mapping (DAMap)} problem in this work.

As shown in  Fig.~\ref{fig.1}b, retraining the final fully connected layers of classifiers using target-domain pseudo-labels is a widely adopted approach\cite{labonte2023towards, qian2024sea} to implementing DAMap. However, it often yields suboptimal performance due to the limited number of trainable parameters. Beyond this simplistic retraining strategy, we can replace these layers with lightweight models for efficient fine-tuning, achieving the dual goals of high efficiency and low training cost (Fig.~\ref{fig.1}c)).

To design an effective DAMap model, we derive insights from the brain’s biological neural networks (BNNs), which can rapidly adapt to new domains and update efficiently without relying on large volumes of labeled data. As Hinton notes \cite{hinton2022forward}, there is no direct evidence for explicit gradient backpropagation in BNNs. Unlike ANNs, which rely on gradient-based function fitting, BNNs adopt a distinct, memory-integrated approach to learning and preserving signal associations. Notably, BNN memory operates through two complementary mechanisms: structurally, it is distributed across complex interconnections of memory-related neural cells \cite{roy2022brain}; functionally, it proceeds via three coordinated learning-memory stages (encoding, storage, and  retrieval) \cite{melton1963implications}. This workflow, dependent on neural plasticity (e.g., synaptic strengthening for memory consolidation), enables the brain to quickly capture, retain, and retrieve signal associations—indirectly supporting its rapid adaptability.

We propose a novel Forward Memorizing Framework, namely MemFlow, to simulate a simplified distributed memorization process, thereby enabling fast \textit{Domain Adaptive Feature Mapping}. It integrates four key design features:
1) Neurons are randomly connected, and signals are transmitted as spikes without continuous activation, which is inspired by spiking activities in the brain \cite{maass1997networks}.
2) It abandons gradient backpropagation; instead, only forward propagation is required to process spiking signals, which are accumulated in the memory units of each neuron and recorded via fuzzy Gaussian distributions to form memory storage.
3) Distributed memories are retrieved and integrated based on confidence scores to generate final classification decisions.
4) It supports reinforced memorization of unlabeled data, enabling efficient adaptation to the target domain.

Extensive experiments conducted on four real-world cross-domain datasets verify that MemFlow can effectively enhance the association between input features and labels in the unlabeled target domain. Specifically, MemFlow achieves up to a 10\% performance improvement while consuming less than 1\% of the computational time required by traditional domain adaptation methods.

The main contributions of this paper are summarized as follows:
\begin{itemize}
	\item We propose a novel gradient-free Forward Memorizing Framework, namely MemFlow, for efficient domain adaptive feature mapping. This network models classification tasks as a distributed memory storage and retrieval process over randomly connected neurons.
	\item We design a spiking information transmission mechanism via neuronal signal accumulation, which enables non-linear signal transformation. We further adopt multiple Gaussian distributions to simplify memory storage and utilize Gaussian-blur-based fuzzy memory to mitigate overfitting.
	
	\item We develop a reinforced memorization mechanism for MemFlow, which supports lightweight and efficient optimization of feature mapping using unlabeled target-domain data.
\end{itemize}
\section{Related works}

\subsection{Domain Adaptation} 
 Domain Adaptation methods\cite{li2024comprehensive} have been extensively investigated for various application scenarios, including object recognition \cite{gopalan2013unsupervised,csurka2017comprehensive,long2018conditional} and semantic segmentation \cite{ hoffman2018cycada, zou2018unsupervised, zhang2019curriculum,toldo2020unsupervised}. Recent research is mainly summarized into two categories: Unsupervised Domain Adaptation(UDA) and Source-free Unsupervised Domain Adaptation(SFUDA).
 
\noindent\textbf{Unsupervised Domain Adaptation} Early Unsupervised Domain Adaptation(UDA)\cite{ganin2015unsupervised,zhang2018collaborative,yang2022unsupervised} approaches often focus on training a learner across domains, which is typically achieved by aligning the diverse cross domain distributions or learning pseudo labels in the target domain. For example, Wang et al. \cite{wang2023cross} achieved domain alignment by minimizing the distance between cross-domain samples. 

\noindent\textbf{Source-free Unsupervised Domain Adaptation} However, due to privacy concerns or limited training conditions, directly accessing source domain data to guide the training process may not be feasible. As a result, related research has focused on tackling source-free unsupervised domain adaptation(SFUDA)\cite{liang2020we, fang2024source}. In particular, Yang et al.\cite{yang2022attracting} optimize an upper-bound objective that enforces prediction consistency among local neighbors in the feature space while promoting divergence among non-neighbors.
 Pan et al.\cite{pan2025overcoming}
propose a prototypical feature compensation network to mitigate feature misalignment by compensating for the target domain's feature discrepancy using source domain prototypes.


All of the above methods require time-consuming optimization of deep features based on gradient back-propagation and are only suitable to be run on the server side. 

\subsection{ANNs without Gradient Back-propagation}
Besides the gradient back-propagation-based neural networks, there are also some gradient-free network structures, such as the Extreme Learning Machine (ELM) and some of their variants \cite{cambria2013extreme,huang2015extreme,tang2015extreme}. ELM adopted the randomized initialized network for the random projection of features and fitted the linear function to these non-linear features. Following a similar idea to ELM, the Broad Learning System (BLS) \cite{chen2017broad} extends the network to support incremental learning for newly added dynamic features. The Echo State Network (ESN) \cite{jaeger2002adaptive} adopted a random projection network to achieve the non-linear features of time series. 
The recently proposed Forward-Forward algorithm \cite{hinton2022forward} attempts to use two forward processes with positive and negative data, respectively, to replace the forward and backward processes in the back-propagation algorithms. 

\textbf{The fundamental difference between MemFlow and the methods described above}. All of the above methods are designed to fit a function that maps the input to the output, while MemFlow memorizes the association between input and output in distributed neurons and is able to perform reinforced memorization in the unlabeled target domain.  


\section{METHODOLOGY}

\subsection{Problem Definition of DAMap}
Most visual classification models can be formulated as the function $c(f(x))$, where $f$ is the feature extractor based on deep models such as ResNet \cite{he2016deep}, and $c$ is the light-weight classifier, such as MLP \cite{rosenblatt1958perceptron,rumelhart1986learning}, that maps the features to output. Given a labeled source domain $D_s = \{(x^s_i,y^s_i) | 0\leq i <n_s\}$ and an unlabeled target domain $D_t = \{ x^t_i |0\leq i <n_t\}$, the \textbf{Domain Adaptive Feature Mapping (DAMap)} aims to transfer the model trained on $D_s$ to $D_t$ and utilize the unlabeled data in $D_t$ to optimize $c$ while freezing the deep feature extractor $f$ (as shown in Fig.~\ref{fig.1}b and Fig.~\ref{fig.1}c. DAMap avoids the time-consuming optimization of deep backbones and is suitable for the lightweight fine-tuning of visual models on edge devices.
\begin{figure*}[tb]
\vspace{-0.5cm}
	\vskip 0.2in
	\centering
	\begin{subfigure}{0.4\linewidth}
		\centering
		\includegraphics[width=\linewidth]{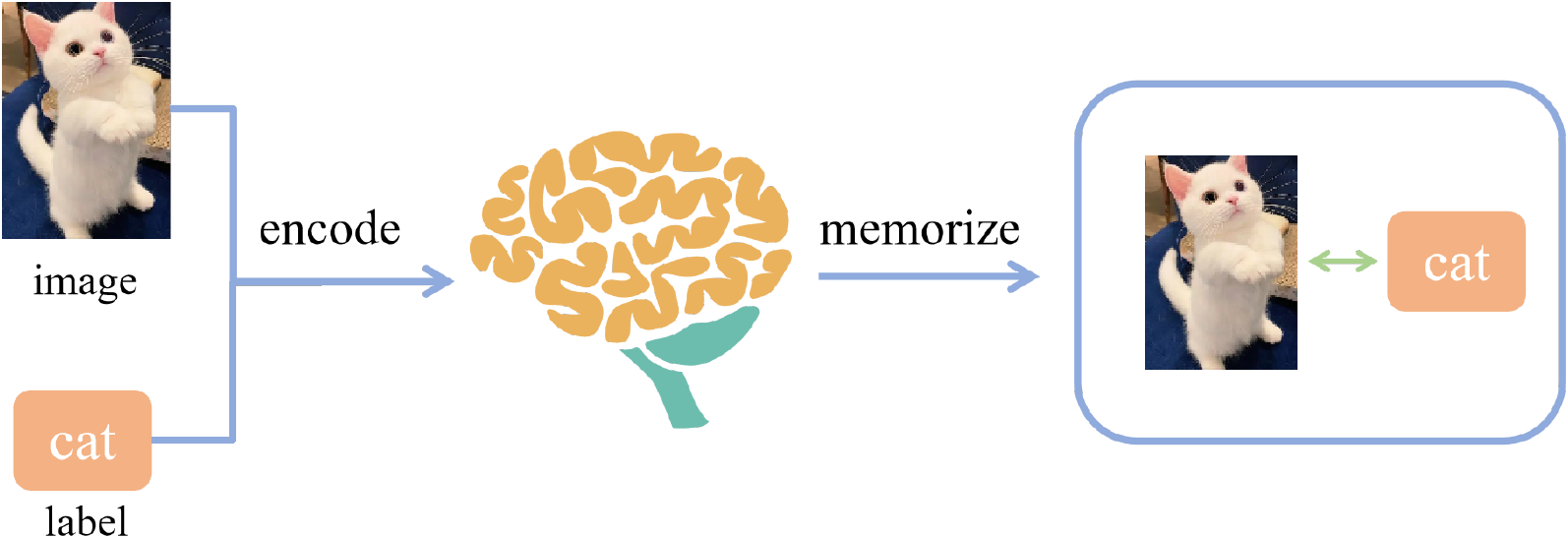}
		\caption{Memory Storage in Brain}
		\label{fig:overview-a}
	\end{subfigure}
	\hspace{20mm}
	\begin{subfigure}{0.4\linewidth}
		\centering
		\includegraphics[width=\linewidth]{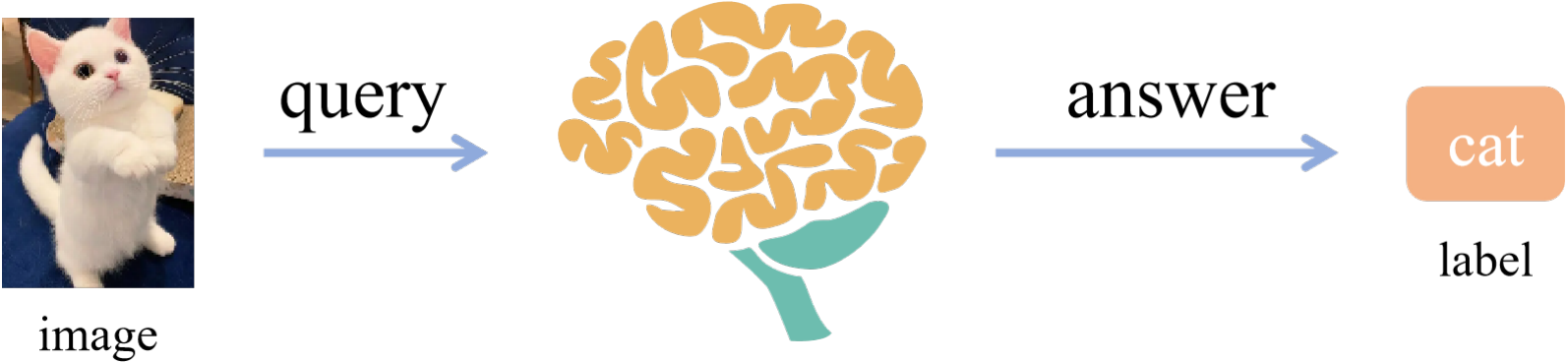}
        \vspace{0.03cm}
		\caption{Memory Retrieval in Brain}
		\label{fig:overview-b}
	\end{subfigure}
	
	\vspace{5mm} 
	
	\begin{subfigure}{0.4\linewidth}
		\centering
		\includegraphics[width=\linewidth]{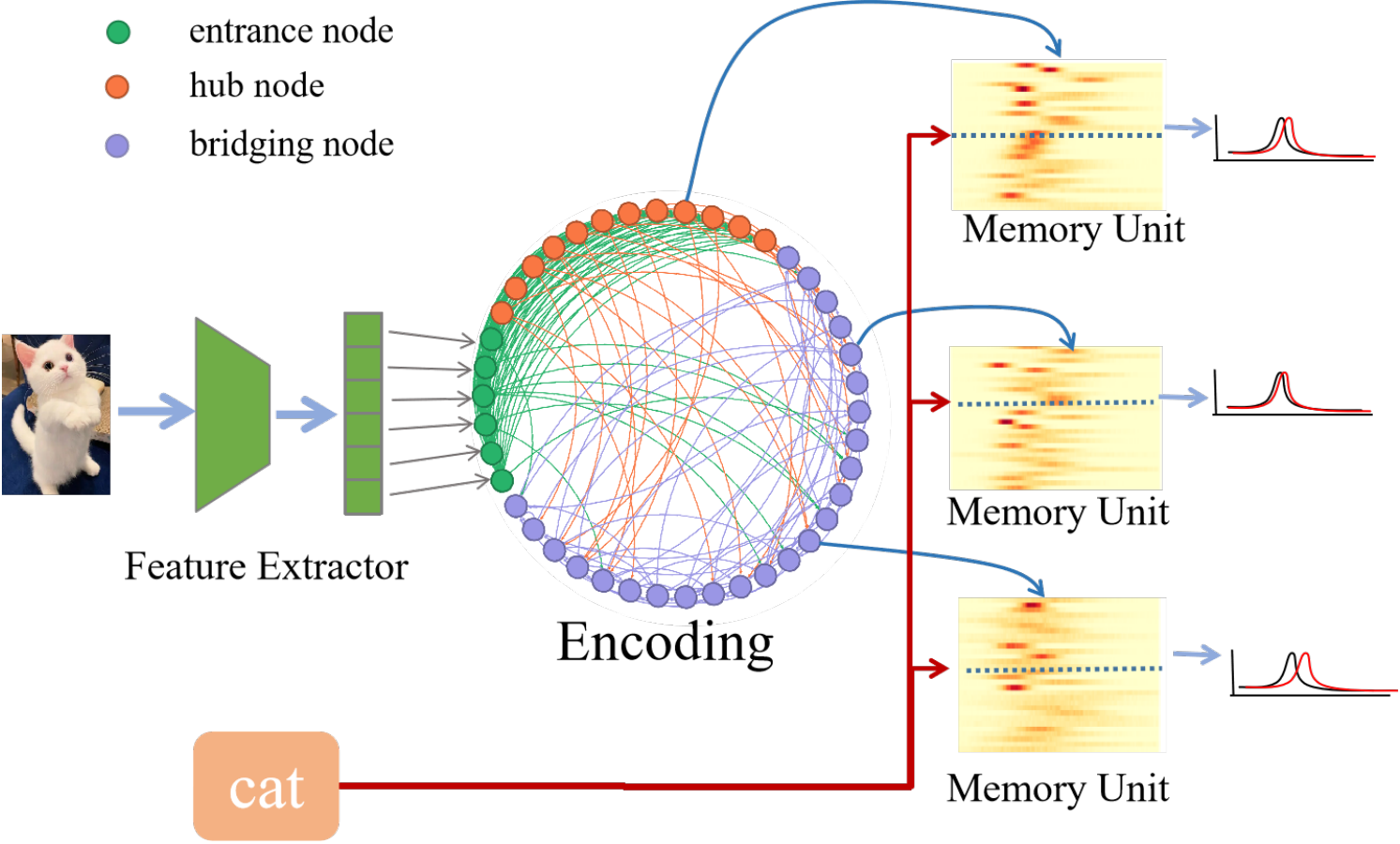}
		\caption{Memory Storage in MemFlow}
		\label{fig:overview-c}
	\end{subfigure}
	\hspace{5mm} 
	\begin{subfigure}{0.55\linewidth}
		\centering
		\includegraphics[width=\linewidth]{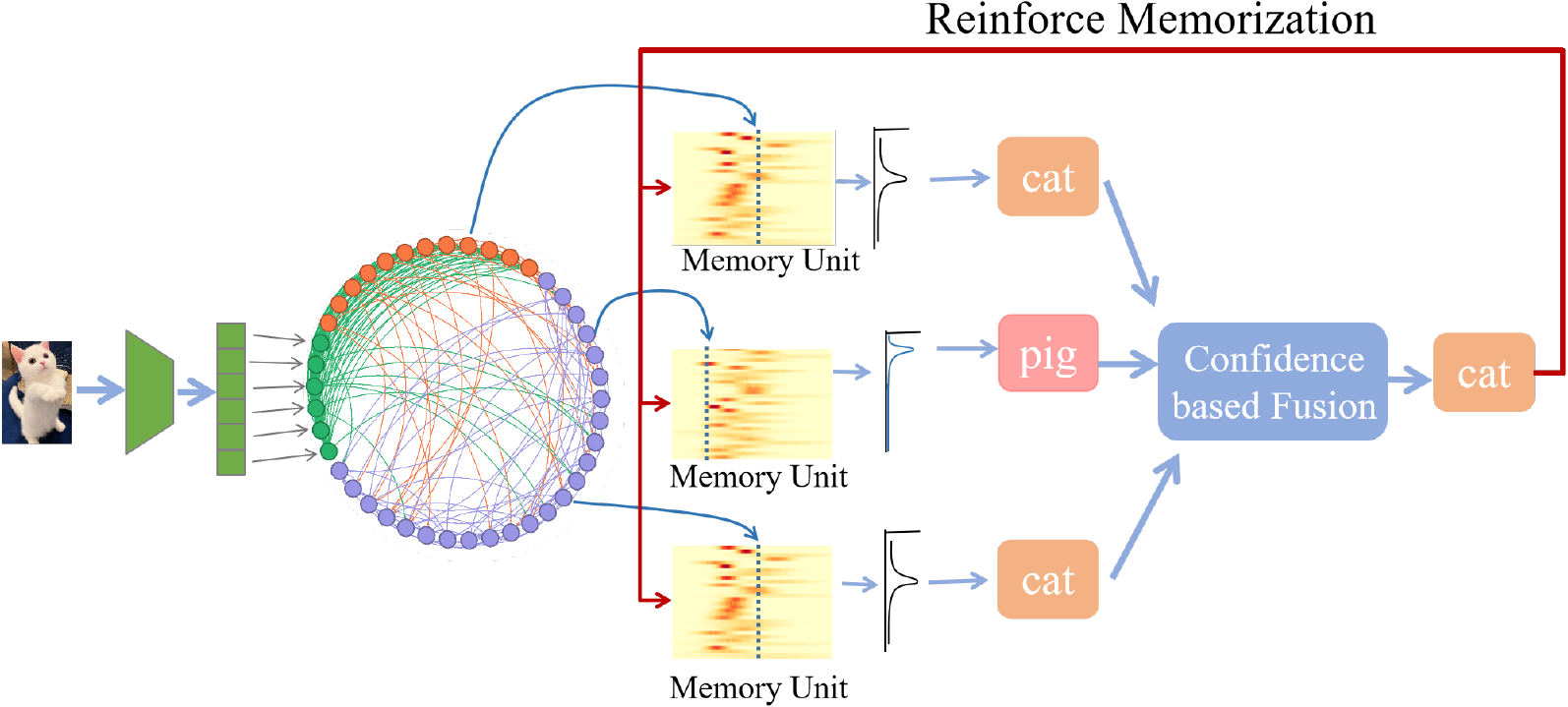}
		\caption{Memory Retrieval in MemFlow}
		\label{fig:overview-d}
	\end{subfigure}
	
	\caption{The framework of MemFlow, including the memory storage and the memory retrieval stages compared with brains.}
	\label{fig:overview}
	\vskip 0.1in
	\vspace{-0.5cm}
\end{figure*}

\subsection{Overview of Forward Memorizing Framework}
As shown in Fig.~\ref{fig:overview}, following the memory storage and retrieval stages of human memorization behaviors, the  Forward Memorizing Framework, namely MemFlow, contains four  memory-based key procedures: 1) \textbf{Memory Encoding} via spiking transmission in randomly connected neural networks achieves a non-linear projection of the input features; 2) \textbf{Distributed Memory Storage}, which is modeled as  Gaussian distributions, records the association between the features and the labels; 3) \textbf{Memory Retrieval} aims to make predictions by integrating the decisions of distributed memory units according to memory confidence; 4) \textbf{Reinforced Memorization}  continuously fine-tunes the memories for domain adaptation according to the predicted labels. The pipeline of MemFlow will be detailed in the following sections.

\subsection{Memory Encoding via Random Projection}


Inspired by the analysis of the brain network structure \cite{bassett2017network}, which contains highly connected hub nodes as well as sparsely linked ones, we adopt a hybrid network topology in MemFlow that includes three types of nodes: entrance nodes, hub nodes, and bridging nodes. As shown in Fig. \ref{fig:overview}, the entrance nodes accept the input features and are densely connected to the hub nodes. The bridging nodes are randomly and sparsely connected to all other nodes. The edges are all directed, and the weights on all edges are randomly initialized in the range $[-1,1]$. 

After the features are fed to the entrance nodes, the signals are propagated through the edges in the network over multiple rounds. Each node accumulates the incoming signals and is activated intermittently to form the spiking output as follows:
\vspace{-0.3cm}
\begin{equation}
	\label{eq3-1-1}
	h_{i,t+1}= h_{i,t} + \sum_{j\in \mathbb{N}_i}o_{j,t} W_{ji}
\end{equation}
\vspace{-0.3cm}
\begin{equation}
	\label{eq3-1-2}
	o_{i,t+1}=
	\left\{
	\begin{array}{ll}
		h_{i,t+1} & if(h_{i,t+1}>0)  \\
		0 &else 
	\end{array}
	\right.
\end{equation}
\vspace{-0.2cm}
\begin{equation}
	\label{eq3-1-3}
	h_{i,t+1}=
	\left\{
	\begin{array}{ll}
		0 & if(h_{i,t+1} > 0)  \\
		h_{i,t+1}  &else 
	\end{array}
	\right.
\end{equation}    
\vspace{-0.2cm}
\begin{equation}
	\label{eq3-1-4}
	m_{i,t+1}=m_{i,t}+o_{i,t+1}
\end{equation}
\vspace{-0.3cm}
\begin{equation}
	\label{eq3-1-5}
	\hat{m}_{i}=m_{i,T}
\end{equation}
Here $h_{i,t}$ and $o_{i,t}(0 \leq i < N)$ indicate the hidden state and output value, respectively, of the  $i^{th}$ node in MemFlow. In the $t^{th} (0 \leq t \leq T)$ round of propagation, the signals collected from the predecessor neighbors are accumulated in the hidden state on each node, as shown in Eq (\ref{eq3-1-1}), where the node $j$ is the predecessor neighbor of the node $i$, and $W_{ji}$ is the weight of the edge from $j$ to $i$. While the hidden state of a node is larger than the threshold $0$, the neuron is activated with the value of the hidden state, as shown in Eq (\ref{eq3-1-2}). Otherwise, the output is $0$ in this round. Once the neuron is activated, the hidden state is cleared as Eq (\ref{eq3-1-3}). The output signal is accumulated as the memory signal $m_{i,t}$ as Eq (\ref{eq3-1-4}), which will be recorded in the memory storage in the future. When an input feature vector $X=(x_1, x_2,...,x_n)$ is fed to the entrance nodes, the output signal $o_{i,0} (1 \leq i \leq n)$ of the entrance nodes is initialized as $x_i$, while $o_{i,0}$ of the other nodes is initialized as zero. Meanwhile, $m_{i,0}$ and $h_{i,0}$ are all set as zero at the beginning.  The steady memory signal of the node $i$ is defined as $
\hat{m}_i$ in Eq (\ref{eq3-1-5}), where $T$ is the maximum number of rounds.

MemFlow differs from traditional spiking neural networks \cite{maass1997networks,Khan2025survey} in two core aspects: (1) its output amplitude is dynamically tied to the magnitude of the positive hidden state (instead of fixed binary spikes), enabling fine-grained encoding of signal strength; (2) it integrates an explicit cumulative memory signal $m_{i,t}$ that transforms transient spiking activity into persistent traces. These designs overcome the limitations of traditional transient, binary signaling, yielding a more robust and specific memory mechanism for the reliable long-term storage and retrieval of input-related information.


\subsection{Distributed Memory Storage}
\label{sec:mem-storage}
Following the distributed memory architecture, each node in MemFlow maintains memory about the association between the  memory signal $\hat{m}$ and the label $y$ for each input instance, as shown in Fig. \ref{fig:overview-c}. The memory unit of the  $i^{th}$ neuron node can be modeled as a two-dimensional image $M_i$, the first dimension of which is the class label $y$, and the other is the memory signal $\hat{m}$. For each incoming feature vector, which is propagated in the network to achieve the memory signal on each neuron by Eq (\ref{eq3-1-5}), the association of the memory signal and the label of the instance can be stored by increasing the value of the corresponding pixel in $M_i$. However, this trivial implementation of memory storage may lead to large consumption of computer memory. 

To reduce the memory cost, we model the memory storage using  probability distributions according to the following theorem:
\begin{theorem}[\textbf{Gaussian Distribution of Neuron Memory}] \label{theorem1}
the distribution of memory signals in the $k^{th}$ class on the $i^{th}$ neuron of MemFlow  follows the Gaussian distribution:
\begin{equation}
\mathbb{P}(\hat{m}_i^k \mid y=k) \sim \mathcal{N}(\mu_i^k, (\sigma_i^k)^2).
 \label{eq6}
\end{equation}
\end{theorem}
The detailed proof is in Section 1 of the supplementary Material. In this way, the storage of memory unit $M_i$ can be represented as $C$ Gaussian distribution $\{\mathcal{N}(\mu_{i}^k, (\sigma_{i}^k)^2)| 1 \leq k \leq C\}$ with  $2*C$ learnable parameters, where C is the number of classes.

While training MemFlow on the labeled source domain, the parameters are incrementally updated in 
batches as follows:
\vspace{-0.2cm}
\begin{equation}
	\mu_{i}^k \leftarrow \beta\mu^k_{i}+(1-\beta)\frac{1}{B}\sum_{b=1}^B \hat{m}_{i,b}^k
	\label{eq7}
\end{equation}
\vspace{-0.3cm}
\begin{equation}
	(\sigma_{i}^k)^2 \leftarrow \beta(\sigma_{i}^k)^2+(1-\beta)\frac{1}{B}\sum_{b=1}^B(\hat{m}_{i,b}^k-\mu_{i}^k)^2
	\label{eq8}
\end{equation}
$\beta$ denotes the temperature parameter between batches, and $B$ denotes the batch size. $\hat{m}_{i,b}^k$ denotes the memory signal received on the $i^{th}$ node while processing the $b^{th}$ instance belonging to the $k^{th}$ class.

Meanwhile, the above memory based learning relies on fitting the probability distribution, so a smaller training set size or an imbalanced label distribution may cause over-fitting of the distribution function on insufficient samples. To increase the generalization ability of the model, we introduce Gaussian blur on the memorized Gaussian distribution (Eq (\ref{eq6})) with the Gaussian kernel function  $g(x_1, x_2)=exp(-\frac{(x_1-x_2)^2}{2(\sigma_1)^2})$ to achieve fuzzy memory:
\vspace{-0.2cm}
\begin{equation}
	\begin{split}
		Q(\hat{m}_i|y=k)=\int_{-\infty}^{+\infty} Pr(\hat{m}|y=k)g(\hat{m},\hat{m}_i)d\hat{m}  \\
		= \frac{\sigma_1}{2\sqrt{2(\sigma_i^k)^2+(\sigma_1)^2}}exp(-\frac{(\hat{m}_i-\mu_i^k)^2}{2(\sigma_i^k)^2+(\sigma_1)^2})
	\end{split}
	\label{eq9}
\end{equation}
The detailed derivation of Eq.(\ref{eq9}) is in Section 3 of the Supplementary Material.

\subsection{Confidence based Memory Retrieval} 
The pipeline of memory retrieval on MemFlow is shown in Fig.~\ref{fig:overview-d}. After the input features of an instance are fed to the entrance nodes and propagated in the network, the $i^{th}$ node will receive the memory signal $\hat{m}_i$ as Eq (\ref{eq3-1-5}) and retrieve its memory unit to gain the conditional probability inference as follows:
\vspace{-0.2cm}
\begin{equation}
	Pr(y=k|\hat{m}_i) = \frac{Q(\hat{m}_i|y=k)}{\sum_{c=1}^C Q(\hat{m}_i|y=c)}
	\label{eq10}
\end{equation}
Here $Q(\hat{m}_i|y=k)$  indicates the fuzzy memory distribution (Eq (\ref{eq9})) achieved at the memory storage stage. The most likely class label predicted by the $i^{th}$ node is 
\begin{equation}
	K_i = \arg\max\limits_k Pr(y=k|\hat{m}_i)
\end{equation}
The confidence of the node to make the prediction can be defined as the likelihood of the memory signal as follows:
\begin{equation}
	c_i = Q(\hat{m}_i|y=K_i)
	\label{eq12}
\end{equation}
The final decision of the whole network is defined as the confidence-based fusion of the predictions of all neurons:
\begin{equation}
	Pr(y=k|X) = \frac{c_i Pr(y=k|\hat{m}_i)}{\sum_j^N c_j}
	\label{eq13}
\end{equation}
The predicted label is as follows: 
\begin{equation}
	\hat{y} = \arg\max\limits_k Pr(y=k|X)
	\label{eq3-4-1}
\end{equation}

\subsection{Reinforced Memorization for Domain Adaptation}

The pairs of samples and pseudo labels $\{(X, \hat{y})\}$ can be fed to MemFlow to conduct reinforced memorizing of the association between the features and labels by updating the parameters according to Eq (\ref{eq7}) and (\ref{eq8}). 

As reported by \cite{litrico2023guiding}, the noise of pseudo labels affects the performance of SFUDA significantly, which may bring the accumulation of errors and make the model over-fit the wrong prediction. To reduce the oscillation of the model caused by the noise, we introduce the confidence $E_i$ of parameter updating on the node $i$ by measuring the likelihood that the node  makes the same prediction as the pseudo label $\hat{y}$:
\begin{equation}
	E_i(\hat{m}_i,\hat{y} ) = Q(\hat{m}_i|y=\hat{y})
\end{equation}
Then the updating of parameters in Eq (\ref{eq7}) and (\ref{eq8}) is rewritten as follows:
\vspace{-0.2cm}
\begin{equation}
	\mu_{i}^{\hat{y}} \leftarrow \beta\mu^{\hat{y}}+(1-\beta)\frac{1}{B}\sum_{b=1}^B E_i(\hat{m}_{i,b}^{\hat{y}},\hat{y}) \hat{m}_{i,b}^{\hat{y}}
	\label{eq16}
\end{equation}
\vspace{-0.2cm}
\begin{equation}
	(\sigma_{i}^{\hat{y}})^2
    \leftarrow \beta(\sigma_{i}^{\hat{y}})^2+(1-\beta)\frac{1}{B}\sum_{b=1}^B E_i(\hat{m}_{i,b}^{\hat{y}},\hat{y}) (\hat{m}_{i,b}^{\hat{y}}-\mu_{i}^{\hat{y}})^2
	\label{eq17}
\end{equation}

The pseudo label with higher confidence on a neuron node will achieve higher weight to update the parameters on the node. After updating the parameters, the model can generate new pseudo labels on the unlabeled data. In this way, the updating can be run in multiple iterations to reinforce the memorization in a self-supervised way, so as to make the learned distribution approach the ground-truth target distribution as shown in Fig. \ref{fig06}.
\begin{table*}[htbp]
	\centering
	\caption{Accuracy(\%) and adaptation time per instance (ms) on the Office-Home dataset, where $\overline{Acc}$ and $\overline{Time}$ are the average accuracy and time cost across all task. $\dagger$ means the reproduced results.}
	\setlength\tabcolsep{3pt} 
	\resizebox{1\linewidth}{!}{
	\begin{tabular}{lcccccccccccccc}
		\toprule
		  & $A\rightarrow C$
		& $A\rightarrow P$        & $A\rightarrow R$        & $C\rightarrow A$        & $C\rightarrow P$        & $C\rightarrow R$        & $P\rightarrow A$        &$P\rightarrow C$       & $P\rightarrow R$        & $R\rightarrow A$        & $R\rightarrow C$        & $R\rightarrow P$ & $\overline{Acc}$($\uparrow$) & $\overline{Time}$($\downarrow$) \\
		\midrule
		\rowcolor{gray!25}
		\multicolumn{15}{l}{\textbf{DAMap Methods}} \\
		\midrule
				w/o DA & 44.6  & 68.5  & 75.1  & 54.1  & 63.5  & 65.6  & 53.2  & 40.3  & 72.5  & 66.2  & 46.9  & 78.6  & 60.8  & - \\
				retrain@last   & 46.1  & 66.8  & 73.2  & 54.6  & 64.2  & 66.3  & 54.4  & 44.0  & 73.8  & 66.5  & 50.1  & 78.9  & 61.6  & 0.082  \\
                retrain@BLS   & 47.0  & 71.6  & 75.9  & 55.3  & 66.2  & 67.6  & 54.6  & 44.3  & 74.1  & 65.8  & 50.2  & 78.6  & 62.6  & 0.088  \\
				retrain@KNN   & 44.2  & 67.1  & 71.6  & 54.4  & 64.7  & 65.6  & 56.4  & 45.5  & 74.6  & 66.8  & 52.4  & 79.2  & 61.9  & 0.305  \\
				retrain@DTC   & 14.4  & 24.2  & 30.6  & 17.3  & 23.4  & 26.8  & 16.9  & 13.1  & 31.6  & 27.3  & 17.6  & 39.2  & 23.5  & 0.556  \\
				retrain@RF   & 40.9  & 60.0  & 68.6  & 48.5  & 58.8  & 60.5  & 47.9  & 39.0  & 69.5  & 63.2  & 46.9  & 76.4  & 56.7  & 2.794  \\
				retrain@XGB   & 36.8  & 54.6  & 63.9  & 36.6  & 48.7  & 49.7  & 32.6  & 29.2  & 56.8  & 50.5  & 35.8  & 66.9  & 46.8  & 3.790  \\
				retrain@SVM   & 47.2  & 69.9  & 74.9  & 52.3  & 63.3  & 64.6  & 51.6  & 42.1  & 72.7  & 65.2  & 48.3  & 77.9  & 60.8  & 2.447  \\
				retrain@BAG   & 45.2  & 67.7  & 72.8  & 52.7  & 65.6  & 65.0  & 56.0 & 45.3  & 74.9  & 67.0  & 52.0  & 78.8  & 61.9  & 0.675  \\
				retrain@NBY   & 48.7  & 72.6  & 77.6  & 52.6  & 67.8  & 67.8  & 50.5  & 40.3  & 72.1  & 60.9  & 47.1  & 76.7  & 61.2  & 0.065  \\
				MemFlow & \textbf{50.4}  & \textbf{76.5}  & \textbf{76.9}  & \textbf{59.3}  & \textbf{71.1}  & \textbf{69.7}  & \textbf{59.9} & \textbf{47.3}  & \textbf{76.5}  & \textbf{69.5}  & \textbf{53.9}  & \textbf{81.0}  & \textbf{66.0}  & \textbf{0.057}  \\
		\midrule
		\rowcolor{gray!25}
		\multicolumn{15}{l}{\textbf{SFUDA Methods}} \\
		SHOT$\dagger$\cite{liang2020we}  & 54.9  & 79.2  & 81.7  & 67.1  & 76.6  & 78.2  & 67.6  & 52.5  & 81.5  & 72.9  & 56.3  & 83.7  & 71.0  & 5.515  \\
		SHOT+MemFlow & \textbf{57.4} & \textbf{79.2} & \textbf{81.8} & \textbf{68.3} & \textbf{78.9} & \textbf{78.8} & \textbf{68.2} & \textbf{55.9} & \textbf{81.8} & \textbf{74.1} & \textbf{59.2} & \textbf{84.6} & \textbf{72.3} & \textbf{6.071}  \\
		AaD$\dagger$\cite{yang2022attracting}   & 56.1  & 77.5  & 80.7  & 67.9  & 80.1  & 79.9  & 67.2  & 57.7  & 82.5  & 72.6  & 59.7  & 85.0  & 72.2  & 2.522  \\
		AaD+MemFlow & \textbf{57.7} & \textbf{78.6} & \textbf{82.4} & \textbf{68.7} & \textbf{80.2} & \textbf{80.0} & \textbf{67.2} & \textbf{57.8} & \textbf{82.9} & \textbf{72.8} & \textbf{59.7} & \textbf{85.5} & \textbf{72.8} & \textbf{5.840}  \\
		PFC$\dagger$\cite{pan2025overcoming}   & 60.4  & 79.8  & 82.0  & 68.6  & 79.2  & 80.2  & 67.8  & 58.8  & 83.1  & 71.0  & 61.5  & 85.6  & 73.2  & 16.570  \\
		PFC+MemFlow & \textbf{60.9} & \textbf{79.9} & \textbf{82.1} & \textbf{68.8} & \textbf{80.1} & \textbf{80.3} & \textbf{67.9} & \textbf{59.4} & \textbf{83.3} & \textbf{73.4} & \textbf{61.8} & \textbf{85.9} & \textbf{73.7} & \textbf{16.609}  \\
		TPDS$\dagger$\cite{tang2024source}  & 60.1  & 80.0  & 82.8  & 70.2  & 78.9  & 80.2  & 69.1  & 56.6  & 82.8  & 74.8  & 61.0  & 85.6  & 73.5  & 20.037  \\
		TPDS+MemFlow & \textbf{60.1} & \textbf{80.5} & \textbf{82.8} & \textbf{70.2} & \textbf{80.3} & \textbf{81.1} & \textbf{69.5} & \textbf{56.7} & \textbf{82.8} & \textbf{75.0} & \textbf{61.4 } & \textbf{85.9} & \textbf{73.9} & \textbf{21.692}  \\
		\bottomrule
	\end{tabular}%
}
	\label{office-home-acc}%
\end{table*}%

The theoretical analysis of the convergence of the reinforced memorization procedure is provided below: 

\begin{theorem}[Convergence of Reinforced Memorization]
Let \(\Theta_t = \{(\mu_i^k(t), \sigma_i^k(t)) \mid 1 \leq i \leq N, 1 \leq k \leq C\}\) denote the memory parameter set of MemFlow at iteration \(t\) of reinforced memorization.  The error rate of pseudo-labels is \(\epsilon \in [0, 1]\). Assume the following conditions hold:
\begin{enumerate}
    \item \textit{Confidence Weight Boundedness:} The update confidence \(E_i(\hat{m}_i, \hat{y}) \in [\underline{E}, \overline{E}]\) for constants \(0 < \underline{E} \leq \overline{E} < 1\).
    \item \textit{Ground-truth Distribution Existence:} There exists a true parameter set \(\Theta^* = \{(\mu_i^{k,*}, \sigma_i^{k,*})\}\) such that \(\mathbb{P}(\hat{m}_i \mid y=k) \sim \mathcal{N}(\mu_i^{k,*}, (\sigma_i^{k,*})^2)\).
\end{enumerate}

Then the parameter sequence \(\{\Theta_t\}\) converges almost surely (a.s.) to a bounded set \(\Theta^\dagger\) satisfying:
\vspace{-0.2cm}
\begin{equation}
    \|\Theta^\dagger - \Theta^*\| \leq O\left(\frac{\epsilon\overline{E}}{1 - \beta }\right),
\end{equation}

where  \(\|\cdot\|\) denotes the Euclidean norm of the parameter vector. Moreover, as \(\epsilon \to 0\), \(\Theta_t\) converges to \(\Theta^*\) at a geometric rate.
\end{theorem}

For \(\epsilon > 0\), the limit error bound is proportional to \(\epsilon\), so higher pseudo-label error slows convergence and expands the error floor. The detailed proof is in Section 2 of the Supplementary Material.

\section{Experiments}
\subsection{Experimental Setup}
\textbf{Datasets}. Comprehensive experiments for DAMap are conducted on the following four popularly used cross-domain datasets: 

\begin{itemize}
	\item Digits: Evaluated on three digit datasets—MNIST\cite{lecun1998gradient} (M), USPS \cite{hull1994database} (U), and SVHN\cite{netzer2011reading} (S)—following the  evaluation protocol of CyCADA\cite{hoffman2018cycada}, including 3 cross-domain adaptation tasks: U$\rightarrow$M, M$\rightarrow$U, and S$\rightarrow$M.
	
	\item Office31: A widely-used domain adaptation dataset with 4,110 images across 31 categories from three domains: Amazon (A), Webcam (W), and DSLR (D). 
	\item Office-Home: A medium-scale benchmark with 65 categories across four domains: Artistic (A), Clip Art (C), Product (P), and Real-World (R). 
	
	\item VisDA-C: A large-scale 12-class recognition dataset with two domains: Synthetic (S) with 152k rendered images and Real (R) with 55k COCO images. Two cross-domain tasks are tested: S$\rightarrow$R and R$\rightarrow$S.
\end{itemize}

\noindent\textbf{Evaluation Metrics}. Following the definition of the DAMap challenge, we train the model on the labeled source domain and optimize it based on the unlabeled data in the target domain while freezing the backbone. The accuracy of classification and timing cost are evaluated in the target domain.

\noindent\textbf{DAMap Baselines}. MemFlow is compared with the following DAMap baselines:
\begin{itemize}
	\item  Retrain@Last. Retrain the last fully connected layers of the classifier as Fig.~\ref{fig.1}b. 
	\item Retrain@$x$, where $x$ can be any lightweight classifier except MemFlow,  indicates the variation model by replacing MemFlow in Fig.~\ref{fig.1}c with another classifier and retraining in the target domain. In particular, 
    retrain@KNN, retrain@DTC, retrain@RF, retrain@SVM and retrain@NBY adopt respectively the K-Nearest Neighbors \cite{cover1967nearest, altman1992introduction},  Decision Tree Classifier \cite{quinlan1987generating}, Random Forest\cite{breiman2001random}, Support Vector Machine \cite{cortes1995support} and Naive Bayes Classifier \cite{hand2001idiot} with  default configuration from \texttt{sklearn}. While retrain@BLS uses the Broad Learning System \cite{chen2017broad} with 500 enhancement nodes. Retrain@XGB uses XGBoost \cite{chen2016xgboost}, in which the number of trees is 100 and the maximum depth of each tree is 6. Retrain@BAG uses the Bagging method \cite{breiman1996bagging} to ensemble 10  KNN base models.
	
\end{itemize}
Furthermore, we use `w/o DA' to indicate transferring the model directly without any further fine-tuning on the target domain.

\noindent\textbf{SFUDA Baselines.}
We also compare MemFlow with the following typical SFUDA methods:
\begin{itemize}
	\item SHOT \cite{liang2020we} adopts self-supervised pseudo-labeling to implicitly align representations from the target domains to the source hypothesis. 
	\item AaD \cite{tang2024source} enforces prediction consistency among local neighbors to enhance transfer in target domain. 
	\item TPDS  \cite{tang2024source}  propose a  target prediction distribution searching  paradigm to overcome the domian shift.
	\item PFC \cite{pan2025overcoming} mitigates feature misalignment by compensating for the target domain's feature discrepancy.
\end{itemize}


\noindent\textbf{Implementation Details}. We utilize the LeNet-5\cite{lecun1998gradient} as the backbones for the simple digital recognition task in the Digits dataset. For the object recognition tasks, we adopt the pre-trained ResNet \cite{he2016deep} models as backbones (ResNet-101 for VisDA-C, and ResNet-50 for  Office-31 and Office-Home), following the experimental configurations in previous works like \cite{deng2019cluster}, \cite{xu2019larger}, and \cite{peng2019moment}.

While training the backbones on the source domains, we adopt mini-batch stochastic gradient descent (SGD) with the momentum as 0.9,  weight decay as $1e^{-3}$, and learning rate $\eta=1e^{-3}$ in VisDA-C  and  $1e^{-2}$ in other datasets. The batch size is set to 64. All DAMap methods are optimized based on frozen features by running 16 epochs over their self-generated pseudo labels, and the best results are recorded. 

In MemFlow, the number of hub nodes and bridging nodes is both 50,  the in-degree of the bridging nodes is 30, and the maximum iterations of propagation $T$ is set to 3 by default. 
\begin{table}[t]
	\centering
	\caption{Accuracy(\%) and adaptation time per instance (ms) on the Digits, Office31 and VisDA-C datasets, where $\overline{Acc}$ and $\overline{Time}$ are the average accuracy and time cost across all task. $\dagger$ means the reproduced results.}
	\resizebox{\linewidth}{!}{
	\begin{tabular}{lcccccc}
		\toprule
		\multicolumn{1}{c}{\multirow{2}[2]{*}{}} & \multicolumn{2}{c}{Digits} & \multicolumn{2}{c}{Office31} & \multicolumn{2}{c}{VisDA-C} \\
		& $\overline{Acc} (\uparrow)$ & $\overline{Time} (\downarrow)$ & $\overline{Acc} (\uparrow)$ & $\overline{Time} (\downarrow)$ & $\overline{Acc} (\uparrow)$ & $\overline{Time} (\downarrow)$ \\
		
		\midrule
		\rowcolor{gray!25}
		\multicolumn{7}{l}{\textbf{DAMap Methods}} \\
		
		w/o DA & 79.6  & -     & 79.0  & -     & 60.0  & - \\
		retrain@last   & 86.9  & 0.063  & 80.4  & 1.397  & 65.0  & 0.096  \\
		retrain@BLS   & 89.0  & 0.044  & 80.4  & 0.084  & 67.2  & 0.045  \\
		retrain@KNN   & 87.1  & 0.155  & 82.3  & 0.046  & 62.7  & 0.210  \\
		retrain@DTC   & 66.6  & 0.458  & 36.2  & 0.382  & 40.1  & 0.602  \\
		retrain@RF   & 84.5  & 1.213  & 77.2  & 1.504  & 57.8  & 2.375  \\
		retrain@X-GB   & 83.4  & 1.324  & 61.4  & 0.636  & 57.0  & 5.695  \\
		retrain@SVM   & 86.6  & 3.720  & 79.6  & 0.473  & 61.1  & 13.235  \\
		retrain@BAG   & 87.8  & 0.944  & 82.6  & 0.753  & 63.2  & 2.018  \\
		retrain@NBY   & 87.5  & 0.022  & 80.8  & 0.058  & 62.5  & 0.026  \\
		MemFlow   & \textbf{89.1} & \textbf{0.013} & \textbf{86.1} & \textbf{0.160} & \textbf{72.1} & \textbf{0.012} \\
		
		\midrule
		\rowcolor{gray!25}
		\multicolumn{7}{l}{\textbf{SFUDA Methods}} \\
		
		SHOT$\dagger$\cite{liang2020we}  & 98.1  & 0.091  & 88.7  & 4.682  & 82.9  & 5.251  \\
		AaD$\dagger$\cite{yang2022attracting}   & 97.7  & 0.319  & 89.9  & 2.656  & 88.0  & 3.060  \\
		PFC$\dagger$\cite{pan2025overcoming}   & 98.1  & 1.056  & 90.5  & 3.901  & 79.1  & 3.766  \\
		TPDS$\dagger$\cite{tang2024source}  & 98.4  & 3.362  & 90.2  & 25.230  & 87.6  &  67.288 
 \\
		
		\bottomrule
	\end{tabular}
    }
\vspace{-0.5cm}
	\label{other-acc}%
\end{table}%

\subsection{Experimental Results}

\textbf{Performance Comparison with DAMap Methods.}  

Table \ref{office-home-acc} reports the accuracy of our proposed MemFlow model across various transfer tasks on the OfficeHome dataset, demonstrating its superior performance over other DAMap methods while  consuming less computational time. Beyond its strong performance on OfficeHome, comprehensive results on the Office31, Digits, and VisDA-C datasets are summarized in Table \ref{other-acc}, where MemFlow consistently attains the highest overall accuracy across all benchmarks. Notably, on the large-scale VisDA-C dataset, MemFlow achieves an average accuracy improvement of over 10\% compared to retrain@last, which only fine-tunes the final layers of the classifier (Fig.~\ref{fig.1}(b)).

\noindent\textbf{Performance Comparison with SFUDA methods.} 

The lower half of Table \ref{office-home-acc} and Table \ref{other-acc} reports MemFlow’s performance under the SFUDA scenario. MemFlow achieves performance comparable to state-of-the-art SFUDA methods while requiring substantially less adaptation time (e.g. less than 1\% of the adaptation time on the large-scale VisDA-C dataset). Notably, MemFlow can also be integrated as a plug-and-play module into existing SFUDA frameworks, with detailed integration configurations provided in Section 4 of the Supplemental Material. Experimental results in Table \ref{office-home-acc} demonstrate that incorporating MemFlow into SFUDA methods can effectively boost model accuracy with minimal additional computational overhead.

\noindent\textbf{Time efficiency.} Experimental results from Table \ref{office-home-acc} and Table \ref{other-acc} demonstrate that MemFlow  has the lowest computational overhead across all evaluated datasets. Notably, MemFlow is substantially more efficient than the lightweight baseline retrain@last, which retrains the final fully connected layers via gradient backpropagation. For instance, on the Office31 dataset, MemFlow’s adaptation time is merely 12.5\% of retrain@last and 0.3\% of PFC\cite{pan2025overcoming}, which is the latest SFUDA method. This result validates the rapid adaptation capability of our proposed forward memorization mechanism. Detailed per-task accuracy comparisons of all datasets are provided in Section 4 of the Supplemental Material.

\begin{table}[t]
	\centering
	\caption{Ablation study about the various components of MemFlow, where $\mathbf{CU}$ indicates the Confidence-based Update, $\mathbf{GB}$ indicates the Gaussian Blur based fuzzy memory and $\mathbf{SM}$ represents the Spiking Mechanism in propagation.}
	\resizebox{\linewidth}{!}{
	\begin{tabular}{cccccccc}
		\toprule
		$\mathbf{CU}$ & $\mathbf{GB}$ & $\mathbf{SM}$ & \multicolumn{1}{l}{Office-31} & \multicolumn{1}{l}{Office-Home} & \multicolumn{1}{l}{Digits} & \multicolumn{1}{l}{VisDA-C} \\
		\midrule
		\ding{55}     & \ding{55}     & \ding{55}     & 73.46  & 42.54  & 60.68  & 35.94  \\
		\ding{51}     & \ding{55}     & \ding{55}     & 84.97  & 55.57  & 64.17  & 42.60  \\
		\ding{55}     & \ding{51}     & \ding{55}     & 84.59  & 65.04  & 89.00  & 70.96  \\
		\ding{51}     & \ding{51}     & \ding{55}     & 85.52  & 65.84  & 89.07  & 71.14  \\
		\ding{51}     & \ding{51}     & \ding{51}     & \textbf{86.08} & \textbf{66.00} & \textbf{89.14} & \textbf{72.08} \\
		\bottomrule
	\end{tabular}%
}
	\label{ablation-study}%
    \vspace{-0.5cm}
\end{table}%
%

\subsection{Ablation study}
To validate the effectiveness of each component of the proposed MemFlow, we conduct ablation studies in this subsection, with the results summarized in Table~\ref{ablation-study}. The experimental results confirm that each module contributes positively to the performance of MemFlow. In particular, the Gaussian Blur (GB) in Eq. (\ref{eq9}) plays a pivotal role, yielding the most significant performance improvement of over 10\% across all datasets. This indicates that the smoothing capability of GB effectively prevents the model's output from being dominated by a single neuron and balances the contribution of each neuron towards a more comprehensive outcome. Second, the Confidence-based Update (CU) in Eq. (\ref{eq16}) and Eq. (\ref{eq17}) incorporates probability-based confidence to guarantee the stable update of neuron nodes, leading to  significant performance. Furthermore, experimental results also demonstrate the advantage of the Spiking Mechanism (SM) in Eq. (\ref{eq3-1-2}) and Eq. (\ref{eq3-1-3}), which introduces a robust and specific memory mechanism for reliable long-term information storage and retrieval.

\subsection{Parameter Analysis }
\begin{figure}[t]
    \vspace{-0.2cm}
	\centering
	\begin{subfigure}[b]{0.49\linewidth}
		\centering
		\includegraphics[width=\linewidth]{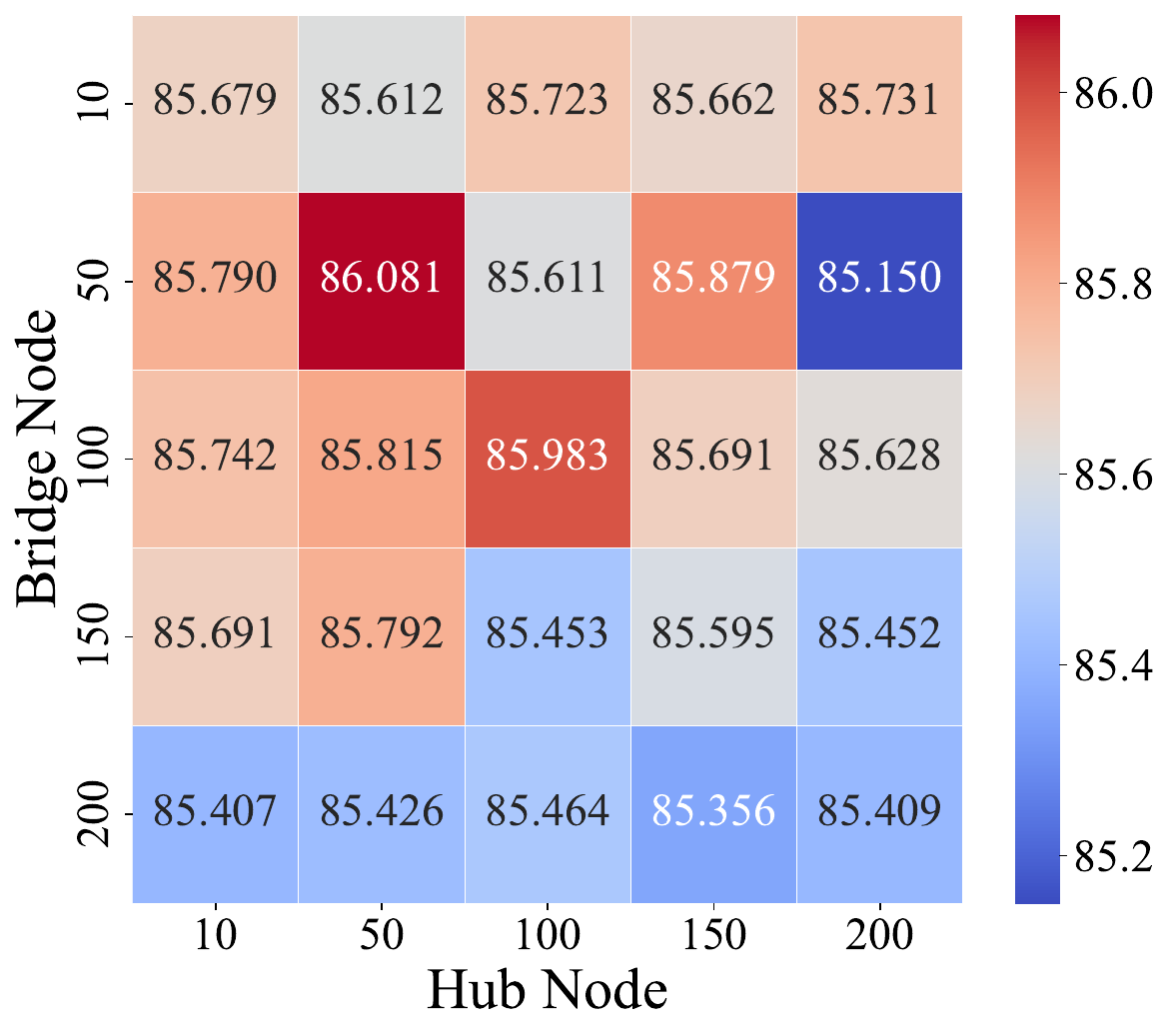} 
		\caption{Office31}
		\label{fig:heatmap_31}
	\end{subfigure}
	\begin{subfigure}[b]{0.49\linewidth}
		\centering
		\includegraphics[width=\linewidth]{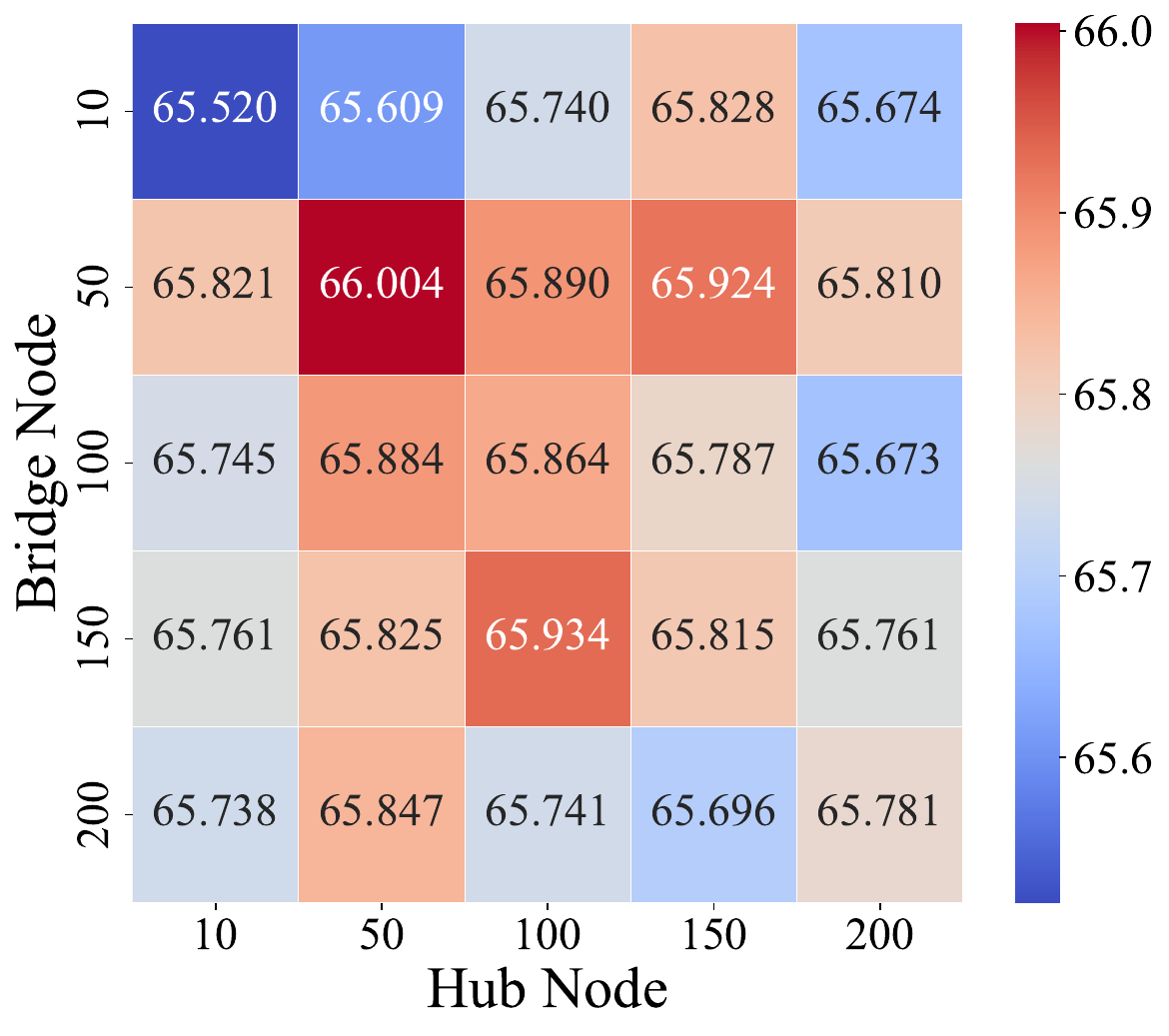} 
		\caption{Office-Home}
		\label{fig:heatmap_oh}
	\end{subfigure}
	\caption{Performance comparison with different Bridge nodes and Hub nodes in the proposed method.}
	\vspace{-0.3cm}
    \label{fig:heatmap}
\end{figure}

\textbf{Bridge/Hub node scale analysis.} In order to investigate the impact of network scale on the performance of domain adaptation, we conducted a sensitivity analysis by systematically varying the number of bridge and hub nodes, as shown in Fig.~\ref{fig:heatmap}.  The results indicate that the performance is generally better when the number of bridge nodes and hub nodes is roughly equal. Further, on the Office-31 dataset, the model performs well when the number of nodes is between 10 and 100; beyond this range, increasing the count yields no further improvement. For the Office-Home dataset, the effective range is approximately 50 to 100 nodes. 

\begin{figure}[t]
	\vskip 0.2in
	\vspace{-0.5cm}
	\begin{center}
		\hspace*{-0.5cm}
		\includegraphics[width=1.08\linewidth]{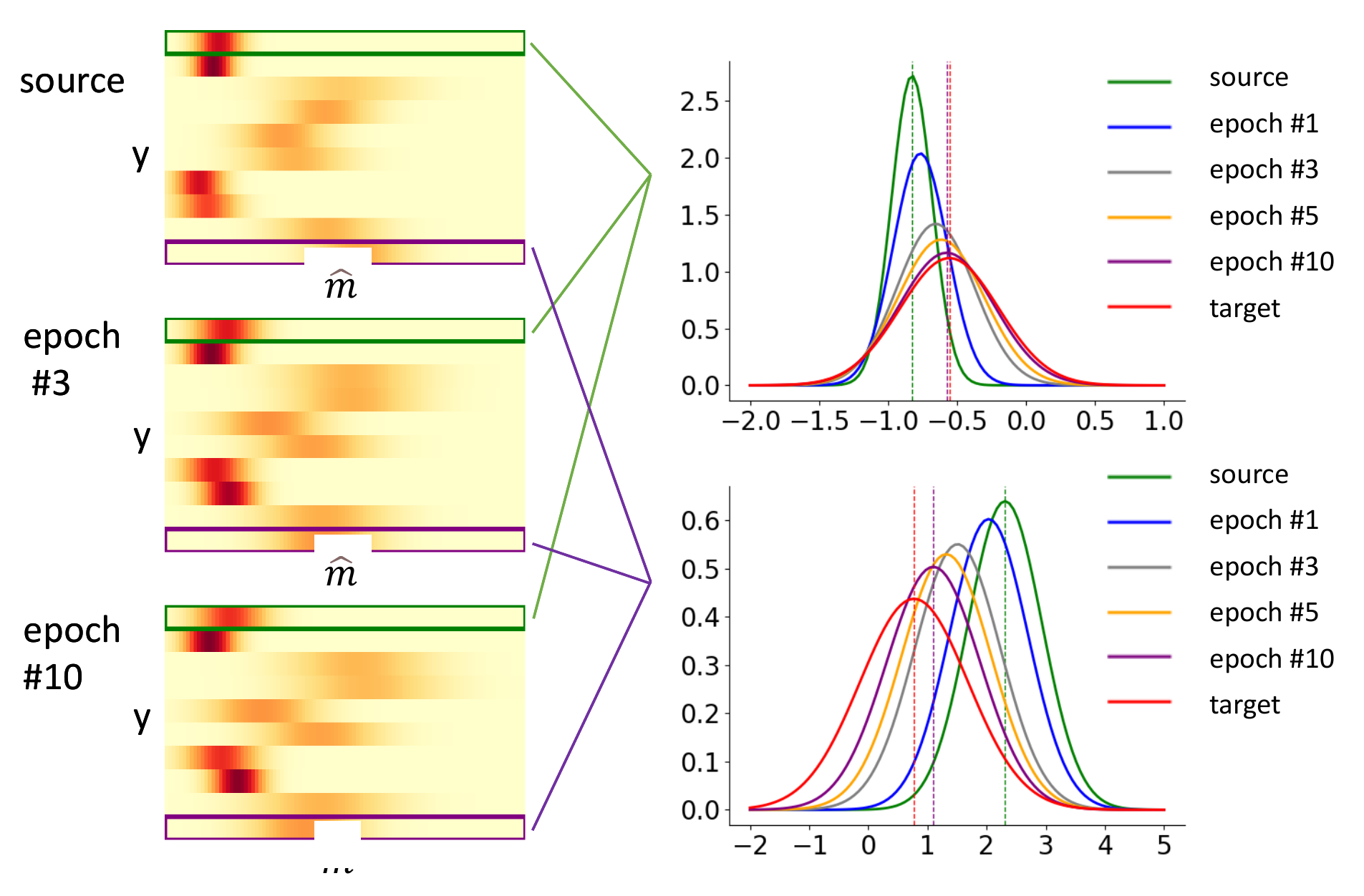}
	\end{center}
	\vspace{-0.4cm}
	\caption{The updating of the memory unit on a neuron while performing the domain adaptation from $M \rightarrow U$ in the Digits dataset. }
	\vskip 0.1in
	\label{fig06}
	\vspace{-1cm}
	
\end{figure}
\begin{figure}[t]
    \vspace{-0.2cm}
	\centering
	\begin{subfigure}[b]{0.49\linewidth}
		\centering
		\includegraphics[width=\linewidth]{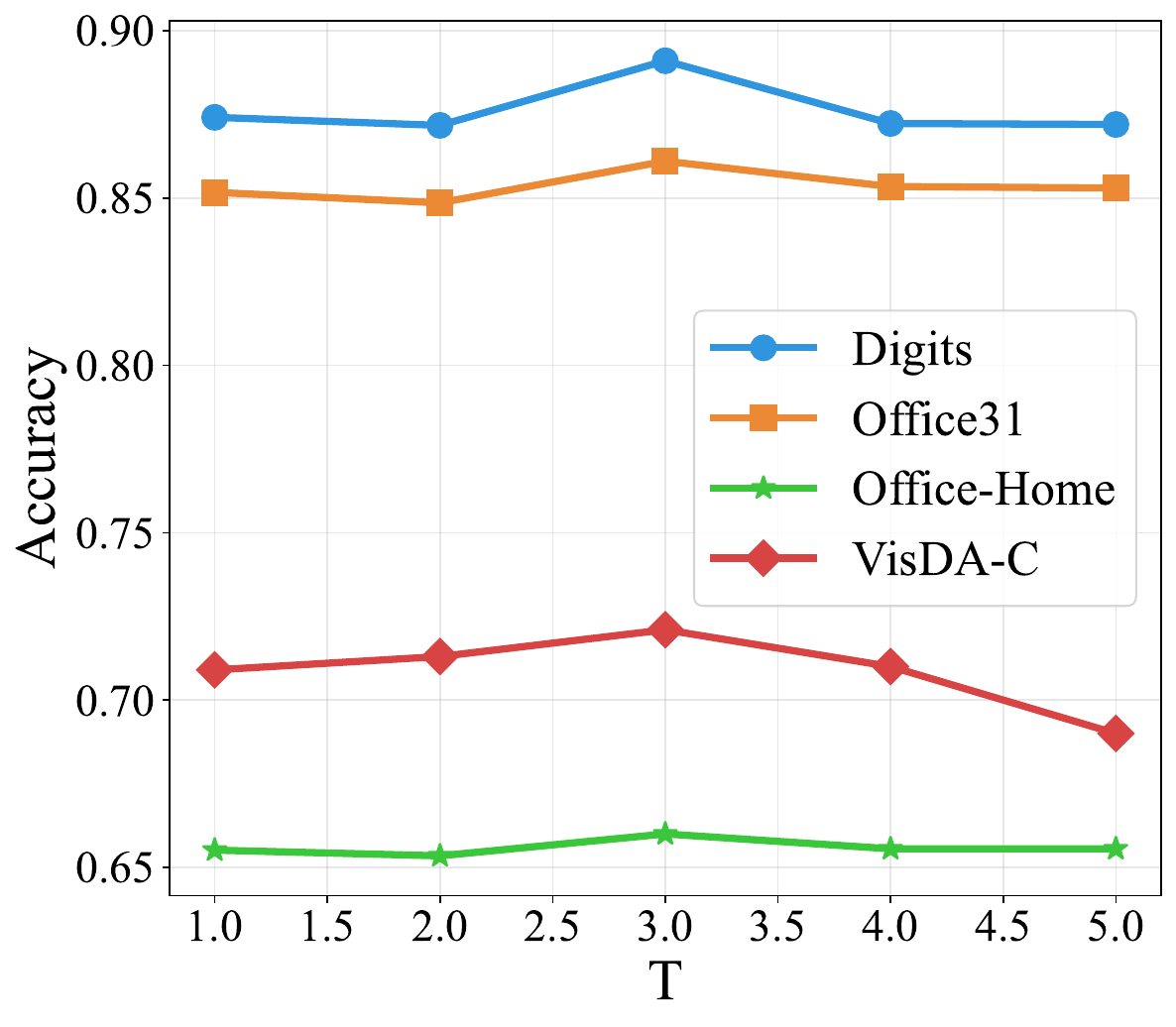} 
		\caption{Numbers of propagation T}
		\label{fig:pa_T}
	\end{subfigure}
	\begin{subfigure}[b]{0.49\linewidth}
		\centering
		\includegraphics[width=\linewidth]{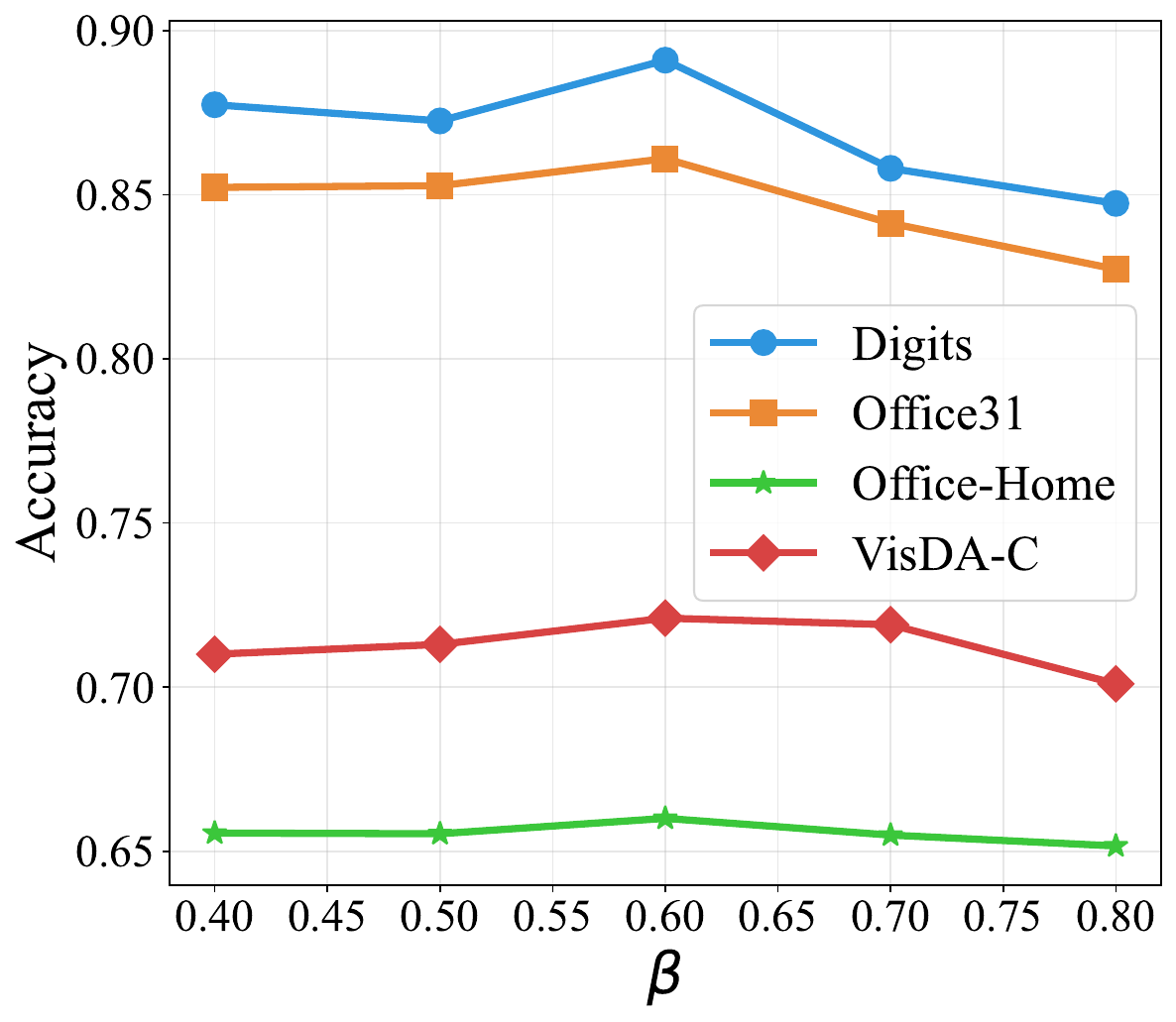} 
		\caption{Memory update rate $\beta$}
		\label{fig:pa_beta}
	\end{subfigure}
	\caption{Performance comparison with different T and $\beta$ in the proposed method.}
    \vspace{-0.7cm}
	\label{fig:pa}
\end{figure}

\noindent\textbf{Hyperparameter Analysis.}
Fig.\ref{fig:pa} illustrates the performance comparison of the MemFlow across various datasets with different parameter values. Fig.~\ref{fig:pa} (a) depicts the performance with respect to the number of propagation steps in MemFlow. The model achieves optimal results across all datasets when the number of propagation steps is set to 3. A larger number of steps leads to a significant decline in accuracy, which we attribute to substantial information loss caused by the spiking mechanism. Conversely, a smaller number of propagation steps results in only minor performance fluctuations, consistent with the regularizing effect of the spiking mechanism on neurons. Fig.~\ref{fig:pa} (b) shows the performance for different memory update rates $\beta$. Similarly, MemFlow attains the best performance when $\beta=0.6$. This can be explained by the fact that excessively high or low update rates may cause overfitting or underfitting in MemFlow, respectively.

\subsection{Visualization}

Fig. \ref{fig06} depicts how the memory units evolve along with the iterations of domain adaptation and highlights the change of the  Gaussian distributions consistent with Eq (\ref{eq7}) and (\ref{eq8}). After multiple rounds of self-supervised learning on pseudo labels, the distributions effectively approach the target distribution, which corresponds to the model supervisedly trained using the labels in the target domain. This validates the remarkable effectiveness of MemFlow in DAMap and is consistent with the convergence property of reinforced memorization, as pointed out in Theorem 2.

\vspace{-0.2cm}

\section{Conclusion}
In this paper, we propose a lightweight Forward Memorizing Framework, namely MemFlow, to support efficient Domain-Adaptive Feature Mapping. In particular, MemFlow learns the association between the features and labels based on the distributed memories in the neurons through spiking information transmission and accumulation. MemFlow is also able to utilize its prediction results on unlabeled data to perform reinforced memorization to support quick domain adaptation. Comprehensive experiments show the superior performance and low timing cost of MemFlow, indicating that MemFlow has good potential for continuous optimization on edge devices.

In the future, we will further explore the relationship between performance and the network topology of MemFlow, inspired by biological neural networks. Meanwhile, we will also extend MemFlow to support more learning tasks, including the multi-modal DAMap problems, to verify the power of distributed memorization. 

\section{Acknowledgements}
This work was supported by the National Key R\&D Program
of China (2023YFA1011601), the Basic and Applied
Basic Research Foundation of Guangdong Province
(2024A1515012287), Science and Technology Key Program of Guangzhou (2023B03J1388).

{
    \small
    \bibliographystyle{ieeenat_fullname}
    \bibliography{main}
}







\maketitlesupplementary

\newcommand{\asconv}{\xrightarrow{\text{a.s.}}}
\section{Supplementary Proof: Gaussian Distribution of Neuron Memory}
\label{sec:suppl_gaussian_proof}

\begin{theorem}[\textbf{Gaussian Distribution of Neuron Memory}] \label{theorem1}
	the distribution of memory signals in the $k^{th}$ class on the $i^{th}$ neuron of MemFlow  follows the Gaussian distribution:
\begin{equation}
\mathbb{P}(\hat{m}_i^k \mid y=k) \sim \mathcal{N}(\mu_i^k, (\sigma_i^k)^2),
\end{equation}. 
\end{theorem}

\begin{lemma}\label{lemma1}
	Define the $t$-th round memory contribution of class $k$ to neuron $i$ as $Z_t^k = o_{i,t}^k$ (where $o_{i,t}^k$ is the neuron's output at round $t$). Then 
	$\{Z_1^k, Z_2^k, \dots, Z_T^k\}$ is i.i.d. across rounds. 
\end{lemma}

\begin{proof}[\textbf{Proof of Lemma \ref{lemma1}}]
	
	For distinct $t_1 \neq t_2$, independence of  $Z_{t_1}^k$ and $Z_{t_2}^k$ holds due to: (i) The activation indicators  are round independent;  (ii) The  signal propagation rules are time-invariant (fixed topology and edge weights); (iii) Input features are i.i.d., so the initial output $o_{j,0}^k = x_j^k$ introduces no round-specific bias. Meanwhile, identical distribution follows from fixed propagation rules, weights, and activation threshold, ensuring $Z_{t_1}^k$ and $Z_{t_2}^k$ share the same distribution.
	
\end{proof}

\begin{proof}[\textbf{Proof of Theorem \ref{theorem1}}] The theorem can be proved in the following steps:
	
	\textit{Step 1: Memory Signal Decomposition}. According to the memory update rule (Eq. (4)), $\hat{m}_i^k = \sum_{t=1}^T o_{i,t}^k = \sum_{t=1}^T Z_t^k$. By Lemma, $\{Z_t^k\}$ is i.i.d. with finite mean $\mu_Z^k$ and positive variance $(\sigma_Z^k)^2$.
	
	\textit{Step 2: Apply Lindeberg-Lévy CLT}. For i.i.d. sequences with finite mean/variance (satisfied by Lemma), the CLT gives:
	\begin{equation}
    \label{eq:CLT}
	\frac{\sum_{t=1}^T Z_t^k - T\mu_Z^k}{\sqrt{T}\sigma_Z^k} \xrightarrow{d} \mathcal{N}(0,1) \quad (T \to \infty). 
	\end{equation}
	
	\textit{Step 3: Gaussian Distribution Deduction}. From Eq. (\ref{eq:CLT}), $\hat{m}_i^k$ converges to $\mathcal{N}(T\mu_Z^k, T(\sigma_Z^k)^2)$. For practical $T$, this approximates to $\mathbb{P}(\hat{m}_i^k \mid y=k) \sim \mathcal{N}(\mu_i^k, (\sigma_i^k)^2)$ with $\mu_i^k = T\mu_Z^k$ and $(\sigma_i^k)^2 = T(\sigma_Z^k)^2$.
	
	\textit{Step 4: Consistency with Gaussian Blur}. The Gaussian blur in Eq. (9) of the main text convolves the memory signal with kernel $g(x_1,x_2) = \exp(-(x_1-x_2)^2/(2\sigma_1^2))$. As convolution preserves Gaussianity for independent Gaussians, we have
	\begin{equation}
	Q(\hat{m}_i^k \mid y=k) = \mathcal{N}(\mu_i^k, (\sigma_i^k)^2 + \sigma_1^2). 
    \end{equation}
	
\end{proof}

\label{sec:conv_proof}

\section{Supplementary Proof: Convergence of Reinforced Memorization}

\begin{theorem}[\textbf{Convergence of Reinforced Memorization}]
\label{theorem2}
	Let \(\Theta_t = \{(\mu_i^k(t), \sigma_i^k(t)) \mid 1 \leq i \leq N, 1 \leq k \leq C\}\) denote the memory parameter set of MemFlow at iteration \(t\) of reinforced memorization.  The error rate of pseudo-labels is \(\epsilon \in [0, 1]\), where \(\epsilon = \mathbb{P}(\hat{y} \neq y^*)\) ( \(y^*\) is the ground-truth label. Assume the following conditions hold:
	\begin{enumerate}
		\item \textit{Confidence Weight Boundedness:} The update confidence \(E_i(\hat{m}_i, \hat{y}) \in [\underline{E}, \overline{E}]\) for constants \(0 < \underline{E} \leq \overline{E} < \infty\).
		\item \textit{Parameter Update Smoothness:} The memory parameters \(\mu_i^k(t), \sigma_i^k(t)\) are updated via Eqs. (16)-(17) with fixed temperature \(\beta \in (0, 1)\) and batch size \(B \geq 1\).
		\item \textit{Ground-truth Distribution Existence:} There exists a true parameter set \(\Theta^* = \{(\mu_i^{k,*}, \sigma_i^{k,*})\}\) such that \(\mathbb{P}(\hat{m}_i \mid y=k) \sim \mathcal{N}(\mu_i^{k,*}, (\sigma_i^{k,*})^2)\).
	\end{enumerate}
	Then the parameter sequence \(\{\Theta_t\}\) converges almost surely (a.s.) to a bounded set \(\Theta^\dagger\) satisfying:
    
    \begin{equation}
		\|\Theta^\dagger - \Theta^*\| \leq O\left(\frac{\epsilon\overline{E}}{1 - \beta}\right),
	\end{equation}

	where  \(\|\cdot\|\) denotes the Euclidean norm of the parameter vector. Moreover, as \(\epsilon \to 0\), \(\Theta_t\) converges to \(\Theta^*\) at a geometric rate.
\end{theorem}

\begin{definition}[Parameter Estimation Error]
	For each neuron \(i\) and class \(k\), define the estimation error of mean and variance at iteration \(t\) as:

    \begin{equation}
		\delta\mu_i^k(t) = \mu_i^k(t) - \mu_i^{k,*}, \quad \delta\sigma_i^k(t) = \sigma_i^k(t) - \sigma_i^{k,*}.
	\end{equation}
	The total parameter error is:
	\begin{equation}
		\Delta(t) = \sum_{i=1}^N \sum_{k=1}^C \left[(\delta\mu_i^k(t))^2 + (\delta\sigma_i^k(t))^2\right].
	\end{equation}
\end{definition}

\begin{lemma}[Bounded Update Noise of  \(\mu_i^k(t)\)]\label{lemma2.1}
	Let \(\tilde{y}_b\) and \(y_b^*\) denote the pseudo-label and ground-truth label of the \(b\)-th sample in a batch respectively.  \(\eta_\mu^k(t)\) denotes the update noise of \(\mu_i^k(t)\) :
    \begin{equation}
	\begin{aligned}
		\eta_\mu^k(t) = & (1 - \beta)\frac{1}{B}\sum_{b=1}^B E_i(\hat{m}_{i,b}, \tilde{y}_b) \mathbb{I}(\tilde{y}_b=k)\hat{m}_{i,b} - \\
		& (1 - \beta)\frac{1}{B}\sum_{b=1}^B E_i(\hat{m}_{i,b}, y_b^*) \mathbb{I}(y_b^*=k)\hat{m}_{i,b},
	\end{aligned}
	\end{equation}
	where \( \mathbb{I}(\cdot)\) is the indicator function.
   Then \(\eta_\mu(t)\) satisfies \(|\eta_\mu(t)| \leq \overline{E} \cdot M \cdot \epsilon\), where \(M = \sup_{b} |\hat{m}_{i,b}|\) (bounded by signal propagation rules).

\end{lemma}
\begin{proof}[\textbf{Proof of Lemma \ref{lemma2.1}}]
	For any sample \(b\),  \(\hat{m}_{i,b}\) is bounded by \(M\) . With \(E_i \leq \overline{E}\), we have:
    \begin{equation}
		\begin{aligned}
			 |\eta_\mu(t)^k| \leq (1 - \beta)\frac{1}{B}\sum_{b=1}^B \overline{E} \cdot |\hat{m}_{i,b}| \cdot \mathbb{I}(\tilde{y}_b=k)\cdot\mathbb{I}(\tilde{y}_b \neq y_b^*) & \\
			\leq \overline{E} \cdot M \cdot \frac{1}{B}\sum_{b=1}^B \mathbb{I}(\tilde{y}_b \neq y_b^*)\asconv \overline{E} \cdot M \cdot \epsilon (as B \to \infty)
		\end{aligned}
	\end{equation}

\end{proof}

\begin{lemma}[Bounded Update Noise of  \(\sigma_i^k(t)\)]\label{lemma2.2}
	Let  \(\eta_\sigma^k(t)\) denotes the update noise of \(\sigma_i^k(t) \) :
 \begin{equation}
	\begin{aligned}
		\eta_\sigma^k(t) = & (1 - \beta)\frac{1}{B}\sum_{b=1}^B E_i(\hat{m}_{i,b}, \tilde{y}_b) \mathbb{I}(\tilde{y}_b = k)(\hat{m}_{i,b} -\mu_i^k)^2- \\
		& (1 - \beta)\frac{1}{B}\sum_{b=1}^B E_i(\hat{m}_{i,b}, y_b^*) \mathbb{I}(y_b^*=k)(\hat{m}_{i,b}-\mu_i^k)^2,
	\end{aligned}
	\end{equation}
 Then \(\eta_\sigma^k(t)\)satisfies  \(|\eta_\sigma^k(t)| \leq \overline{E} \cdot M' \cdot \epsilon\)  , where  \(M' = \sup_{b} (\hat{m}_{i,b} - \mu_i^k)^2\) .
\end{lemma}

\begin{proof}[\textbf{Proof of Lemma \ref{lemma2.2}}]
For any sample b, \((\hat{m}_{i,b} - \mu_i^k)^2\leq M'\). With \(E_i \leq \overline{E}\), we have: 
\begin{equation}
		\begin{aligned}
			 |\eta_\sigma(t)^k| \leq (1 - \beta)\frac{1}{B}\sum_{b=1}^B \overline{E} \cdot M' \cdot \mathbb{I}(\tilde{y}_b=k)\cdot\mathbb{I}(\tilde{y}_b \neq y_b^*) & \\
			\leq \overline{E} \cdot M' \cdot \frac{1}{B}\sum_{b=1}^B \mathbb{I}(\tilde{y}_b \neq y_b^*)\asconv \overline{E} \cdot M' \cdot \epsilon (as B \to \infty)
		\end{aligned}
	\end{equation}
\end{proof}

\begin{lemma}[Contraction of Error Expectation for Multi-Class]\label{lemma2.3}
	The expected total error satisfies:
    \begin{equation}
		\mathbb{E}[\Delta(t+1)] \leq \beta^2 \mathbb{E}[\Delta(t)] + K \cdot \epsilon^2,
	\end{equation}

	where \(K = N \cdot C \cdot \max\left\{(\overline{E}M)^2, (\overline{E}M')^2\right\} \cdot  \frac{1+\beta}{1 - \beta}\) (constant), \(N\) is the number of neurons, and other variables are defined as in Lemma \ref{lemma2.1} and \ref{lemma2.2}.
\end{lemma}
\begin{proof}[\textbf{Proof of Lemma \ref{lemma2.3}}]
	We first analyze the mean parameter error, then extend to variance error.
	
	\noindent\textbf{Step 1: Mean Parameter Error Recurrence.}
	From the update rule Eq. (16), we have:
    \begin{equation}
    \small
		\mu_i^k(t+1) = \beta \mu_i^k(t) + (1 - \beta)\frac{1}{B}\sum_{b=1}^B E_i(\hat{m}_{i,b}, \tilde{y}_b)\mathbb{I}(\tilde{y}_b=k) \hat{m}_{i,b}.
	\end{equation}
	
	Substitute \(\mu_i^k(t) = \mu_i^{k,*} + \delta\mu_i^k(t)\) and use the stationary condition of true parameters:
    \begin{equation}
		\mu_i^{k,*} = \beta \mu_i^{k,*} + (1 - \beta)\frac{1}{B}\sum_{b=1}^B E_i(\hat{m}_{i,b}, y_b^*) \mathbb{I}(y_b^*=k)\hat{m}_{i,b}.
	\end{equation}
	
	Subtracting these two equations gives the error recurrence:
    \begin{equation}
		\delta\mu_i^k(t+1) = \beta \delta\mu_i^k(t) + \eta_\mu^k(t),
	\end{equation}

	where \(\eta_\mu(t)\) is the update noise from Lemma \ref{lemma2.1}. Taking the square and expectation:
    \begin{equation}
		\begin{aligned}
			\mathbb{E}\left[(\delta\mu_i^k(t+1))^2\right] = \beta^2 \mathbb{E}\left[(\delta\mu_i^k(t))^2\right] + \\
			2\beta \mathbb{E}\left[\delta\mu_i^k(t) \cdot \eta_\mu^k(t)\right] + \mathbb{E}\left[(\eta_\mu^k(t))^2\right].
		\end{aligned}
        \label{eq.33}
	\end{equation}

	\noindent\textbf{Step 2: Bounding Cross and Noise Terms.}
	By Cauchy-Schwarz inequality:
    \begin{equation}
		\left|\mathbb{E}\left[\delta\mu_i^k(t) \cdot \eta_\mu^k(t)\right]\right| \leq \sqrt{\mathbb{E}\left[(\delta\mu_i^k(t))^2\right]} \cdot \sqrt{\mathbb{E}\left[(\eta_\mu^k(t))^2\right]}.
	\end{equation}

	From Lemma \ref{lemma2.1}, \(\mathbb{E}\left[(\eta_\mu(t))^2\right] \leq \left(\overline{E}M \cdot \epsilon\right)^2 \). 
    We prove \(\mathbb{E}\left[(\delta\mu_i^k(t))^2\right] \leq D\) (uniformly bounded) by induction. Firstly, in the base case (\(t=0\)), initial parameters are finite, so \(\mathbb{E}\left[(\delta\mu_i^k(0))^2\right] \leq D_0\) (constant).
	In the inductive step, assume \(\mathbb{E}\left[(\delta\mu_i^k(t))^2\right] \leq D\), then:

    \begin{equation}
		\mathbb{E}\left[(\delta\mu_i^k(t+1))^2\right] \leq \beta^2 D + 2\beta \sqrt{D C_\mu} + C_\mu.
	\end{equation},where \( C_\mu\ =(\overline{E}M \cdot \epsilon)^2 \) . 
	Setting \(D = \frac{C_\mu}{(1 - \beta)^2}\) (positive solution) satisfies the inequality, so \(\sqrt{\mathbb{E}\left[(\delta\mu_i^k(t))^2\right]} \leq \frac{\sqrt{C_\mu}}{1 - \beta}\).
	
	Substituting back into Eq.(\ref{eq.33}), we have:
    \begin{equation}
		\begin{aligned}
			\mathbb{E}\left[(\delta\mu_i^k(t+1))^2\right] & \leq \beta^2 \mathbb{E}\left[(\delta\mu_i^k(t))^2\right] + 2\beta \cdot \frac{C_\mu}{1 - \beta} + C_\mu  \\
			& = \beta^2 \mathbb{E}\left[(\delta\mu_i^k(t))^2\right] + \frac{1+
				\beta}{1 - \beta}.C_\mu
		\end{aligned}
	\end{equation}

	\noindent\textbf{Step 3: Extending to Variance and Total Error.}
	For variance error \(\delta\sigma_i^k(t)\), the same derivation gives:
    \begin{equation}
		\mathbb{E}\left[(\delta\sigma_i^k(t+1))^2\right] \leq \beta^2 \mathbb{E}\left[(\delta\sigma_i^k(t))^2\right] + \frac{1+\beta}{1 - \beta}C_\sigma,
	\end{equation}
	where \(C_\sigma = (\overline{E}M' \cdot \epsilon)^2\). Summing over all \(i\) and \(k\):
    
		\[\mathbb{E}[\Delta(t+1)] \leq \beta^2 \mathbb{E}[\Delta(t)] + N C \cdot  \frac{1+\beta}{1 - \beta} \cdot\max\{C_\mu, C_\sigma\}.\]

	Defining \(K = N \cdot C \cdot \max\left\{(\overline{E}M)^2, (\overline{E}M')^2\right\} \cdot  \frac{1+\beta}{1 - \beta}\)  completes the proof.
\end{proof}

\begin{proof}[\textbf{Proof of Theorem \ref{theorem2}}]
	We proceed in two key steps to establish convergence and error bounds.
	
	\noindent\textbf{Step 1: Solving the Error Recurrence.}
	From Lemma \ref{lemma2.3}, the expected total error follows a contractive linear recurrence:
    \begin{equation}
		\mathbb{E}[\Delta(t+1)] \leq \beta^2 \mathbb{E}[\Delta(t)] + K \cdot \epsilon^2.
	\end{equation}
	
	Unfolding the recurrence (summing the geometric series):
    \begin{equation}
		\mathbb{E}[\Delta(t)] \leq \beta^{2t} \Delta(0) + K \cdot \epsilon^2 \sum_{s=0}^{t-1} (\beta^2)^s.
	\end{equation}
	
	Since \(\beta \in (0, 1)\), the geometric series converges to  \\ \(\sum_{s=0}^\infty (\beta^2)^s = \frac{1}{1 - \beta^2}\). Thus:
    \begin{equation}
		\mathbb{E}[\Delta(t)] \leq \beta^{2t} \Delta(0) + \frac{K \epsilon^2}{ (1 - \beta^2)}.
	\end{equation}
	
	As \(t \to \infty\), \(\beta^{2t} \to 0\), so \(\mathbb{E}[\Delta(t)]\) converges to the bounded value \(\frac{K \epsilon^2}{ (1 - \beta^2)}\).
	
	\noindent\textbf{Step 2: Almost Sure Convergence.}
	By Markov's inequality, for any \(\delta > 0\):
	
    \begin{equation}
		\mathbb{P}(\Delta(t) > \delta) \leq \frac{\mathbb{E}[\Delta(t)]}{\delta} \to \frac{K \epsilon^2}{\delta  (1 - \beta^2)}.
	\end{equation}

	Choosing \(\delta > \frac{K \epsilon^2}{ (1 - \beta^2)}\) implies \(\mathbb{P}(\Delta(t) > \delta) \to 0\) as \(t \to \infty\). By the Borel-Cantelli lemma, the events \(\{\Delta(t) > \delta\}\) occur only finitely often a.s., so \(\Delta(t)\) converges a.s. to a set \(\Theta^\dagger\) with:
    \begin{equation}
		\|\Theta^\dagger - \Theta^*\| = \sqrt{\Delta^\dagger} \leq \sqrt{\frac{K \epsilon^2}{(1 - \beta^2)}} = O\left(\frac{\epsilon\overline{E}}{1 - \beta}\right).
	\end{equation}

	\noindent\textbf{Convergence Rate}
	For \(\epsilon = 0\) (perfect pseudo-labels), \(\eta_\mu(t) = \eta_\sigma(t) = 0\), so \(\Delta(t) = \beta^{2t} \Delta(0)\) (geometric convergence). For \(\epsilon > 0\), the limit error bound is proportional to \(\epsilon\), so higher pseudo-label error slows convergence and expands the error floor.

\end{proof}

\section{Derivation of Equation (9)}
We present a detailed derivation of the fuzzy memory distribution $Q(\hat{m}_i|y=k)$ in Equation (9), which is obtained by convolving the memory signal distribution $Pr(\hat{m}|y=k)$ with the Gaussian blur kernel $g(\hat{m},\hat{m}_i)$. For brevity, we use $\hat{\sigma}_i^k$ to denote$(\sigma_i^k)^2$, and use $\hat{\sigma}_1$ to denote $({\sigma}_1)^2$

\subsection{Preliminaries and Notations}
We first clarify the mathematical formulations of key components involved in the derivation:

\begin{enumerate}
	\item \textit{Memory signal distribution:} For the $i$-th neuron and $k$-th class, the memory signal $\hat{m}$ follows a Gaussian distribution $N(\mu_i^k, \hat{\sigma_i}^k)$, where $\mu_i^k$ and $\hat{\sigma_i}^k$ denote the mean and variance, respectively:
    \begin{equation}
		Pr(\hat{m}|y=k) = \frac{1}{\sqrt{2\pi \hat{\sigma_i}^k}} \exp\left( -\frac{(\hat{m} - \mu_i^k)^2}{2\hat{\sigma_i}^k} \right)
	\end{equation}

	\item \textit{Gaussian blur kernel:}  The kernel function for smoothing memory signals is defined as (without normalization factor, consistent with the main text):
    \begin{equation}
		g(\hat{m}, \hat{m}_i) = \exp\left( -\frac{(\hat{m} - \hat{m}_i)^2}{2\hat{\sigma}_1} \right)
	\end{equation}
	
	where $\hat{\sigma}_1$ is the bandwidth parameter of the Gaussian kernel, and $\hat{m}_i$ is the memory signal of the $i$-th neuron for the current input.
	\item \textit{Convolution integral definition:} 
	The fuzzy memory distribution $Q(\hat{m}_i|y=k)$ is formally defined as the convolution of $Pr(\hat{m}|y=k)$ and $g(\hat{m},\hat{m}_i)$:
    \begin{equation}
		Q(\hat{m}_i|y=k) = \int_{-\infty}^{+\infty} Pr(\hat{m}|y=k) \cdot g(\hat{m}, \hat{m}_i) d\hat{m}
	\end{equation}

\end{enumerate}

\subsection{Step 1: Substitute and Combine the Integrand}
Substitute $Pr(\hat{m}|y=k)$ and $g(\hat{m},\hat{m}_i)$ into the convolution integral, and extract constant terms outside the integral sign:
\begin{equation}
	\begin{aligned}
		Q(\hat{m}_i|y=k) & = \frac{1}{\sqrt{2\pi \hat{\sigma_i}^k}} \int_{-\infty}^{+\infty} \exp( -\frac{(\hat{m} - \mu_i^k)^2}{2\hat{\sigma_i}^k}  \\
		& -\frac{(\hat{m} - \hat{m}_i)^2}{2\hat{\sigma}_1}) d\hat{m}
	\end{aligned}
\end{equation}

For brevity, we denote the exponent term as $E(\hat{m})$:
\begin{equation}
	E(\hat{m}) = -\frac{1}{2} \left[ \frac{(\hat{m} - \mu_i^k)^2}{\hat{\sigma_i}^k} + \frac{(\hat{m} - \hat{m}_i)^2}{\hat{\sigma}_1} \right]
\end{equation}

\subsection{Step 2: Complete the Square for the Exponent Term}
Expand the quadratic terms in $E(\hat{m})$:
\begin{equation}
	\begin{aligned}
		& (\hat{m} - \mu_i^k)^2 = \hat{m}^2 - 2\mu_i^k \hat{m} + (\mu_i^k)^2, \quad \\
		& (\hat{m} - \hat{m}_i)^2 = \hat{m}^2 - 2\hat{m}_i \hat{m} + (\hat{m}_i)^2
	\end{aligned}
\end{equation}

Substitute these expansions back into $E(\hat{m})$ and group like terms (i.e., $\hat{m}^2$, $\hat{m}$, and constant terms):
\begin{equation}
	\begin{aligned}
		E(\hat{m}) = -\frac{1}{2} & [ \hat{m}^2 \left( \frac{1}{\hat{\sigma_i}^k} + \frac{1}{\hat{\sigma}_1} \right) - 2\hat{m} \left( \frac{\mu_i^k}{\hat{\sigma_i}^k} + \frac{\hat{m}_i}{\hat{\sigma}_1} \right) \\
		& + \left( \frac{(\mu_i^k)^2}{\hat{\sigma_i}^k} + \frac{(\hat{m}_i)^2}{\hat{\sigma}_1} \right)
	\end{aligned}
\end{equation}

We define simplified coefficients to streamline the completing-the-square process:
Quadratic coefficient: $a = \frac{1}{\hat{\sigma_i}^k} + \frac{1}{\hat{\sigma}_1} = \frac{\hat{\sigma_i}^k + \hat{\sigma}_1}{\hat{\sigma_i}^k \hat{\sigma}_1}$, 
Linear coefficient: $b = \frac{\mu_i^k}{\hat{\sigma_i}^k} + \frac{\hat{m}_i}{\hat{\sigma}_1}$,
and Constant term: $c = \frac{(\mu_i^k)^2}{\hat{\sigma_i}^k} + \frac{(\hat{m}_i)^2}{\hat{\sigma}_1}$

Applying the completing-the-square method to the quadratic expression inside the brackets:
\begin{equation}
	a \hat{m}^2 - 2b \hat{m} + c = a \left( \hat{m} - \frac{b}{a} \right)^2 + \left( c - \frac{b^2}{a} \right)
\end{equation}

Substitute this back into $E(\hat{m})$:
\begin{equation}
	E(\hat{m}) = -\frac{a}{2} \left( \hat{m} - \frac{b}{a} \right)^2 - \frac{1}{2} \left( c - \frac{b^2}{a} \right)
\end{equation}

\subsection{Step 3: Evaluate the Gaussian Integral}
Substitute the simplified $E(\hat{m})$ back into the convolution integral, and split the exponent into two separate terms (with the constant part moved outside the integral):
\begin{equation}
	\begin{aligned}
		Q(\hat{m}_i|y=k) = \frac{1}{\sqrt{2\pi \hat{\sigma_i}^k}} \exp\left( -\frac{1}{2} \left( c - \frac{b^2}{a} \right) \right) \\
		\cdot\int_{-\infty}^{+\infty} \exp\left( -\frac{a}{2} \left( \hat{m} - \frac{b}{a} \right)^2 \right) d\hat{m}
	\end{aligned}
\end{equation}

We use the standard Gaussian integral formula \\ $\int_{-\infty}^{+\infty} \exp(-p x^2) dx = \sqrt{\frac{\pi}{p}}$ (by substituting $x = \hat{m} - \frac{b}{a}$ and $p = \frac{a}{2}$):
\begin{equation}
	\int_{-\infty}^{+\infty} \exp\left( -\frac{a}{2} x^2 \right) dx = \sqrt{\frac{2\pi}{a}}
\end{equation}

Substitute the integral result into the expression and simplify the constant term:
\begin{equation}
	Q(\hat{m}_i|y=k) = \frac{1}{\sqrt{2\pi \hat{\sigma_i}^k}} \exp\left( -\frac{1}{2} \left( c - \frac{b^2}{a} \right) \right) \cdot \sqrt{\frac{2\pi}{a}}
\end{equation}

Cancel out $\sqrt{2\pi}$ and further simplify:
\begin{equation}
	Q(\hat{m}_i|y=k) = \frac{1}{\sqrt{a \hat{\sigma_i}^k}} \exp\left( -\frac{1}{2} \left( c - \frac{b^2}{a} \right) \right)
\end{equation}

\subsection{Step 4: Simplify the Constant Term in the Exponent}
Substitute $a, b, c$ back into $c - \frac{b^2}{a}$ and simplify. First, expand $b^2$:
\begin{equation}
	\begin{aligned}
		& b^2  = \left( \frac{\mu_i^k}{\hat{\sigma_i}^k} + \frac{\hat{m}_i}{\hat{\sigma}_1} \right)^2 \\
		& = \frac{(\mu_i^k)^2 \hat{\sigma}_1^2 + 2\mu_i^k \hat{m}_i \hat{\sigma_i}^k \hat{\sigma}_1 + (\hat{m}_i)^2 \sigma_i^{k^2}}{\sigma_i^{k^2} \hat{\sigma}_1^2}
	\end{aligned}
\end{equation}

Substitute $a = \frac{\hat{\sigma_i}^k + \hat{\sigma}_1}{\hat{\sigma_i}^k \hat{\sigma}_1}$ and simplify the fraction $\frac{b^2}{a}$:
\begin{equation}
	\frac{b^2}{a} = \frac{(\mu_i^k)^2 \hat{\sigma}_1 + 2\mu_i^k \hat{m}_i \hat{\sigma_i}^k + (\hat{m}_i)^2 \hat{\sigma_i}^k}{\hat{\sigma_i}^k + \hat{\sigma}_1}
\end{equation}

Combine this with $c$ and simplify (noting that cross terms cancel out):
\begin{equation}
	c - \frac{b^2}{a} = \frac{(\hat{m}_i - \mu_i^k)^2}{\hat{\sigma_i}^k + \hat{\sigma}_1}
\end{equation}

\subsection{Step 5: Final Simplification}
Substitute $a = \frac{\hat{\sigma_i}^k + \hat{\sigma}_1}{\hat{\sigma_i}^k \hat{\sigma}_1}$ into the constant term:
\begin{equation}
	\frac{1}{\sqrt{a \hat{\sigma_i}^k}} = \frac{1}{\sqrt{\frac{\hat{\sigma_i}^k + \hat{\sigma}_1}{\hat{\sigma_i}^k \hat{\sigma}_1} \cdot \hat{\sigma_i}^k}} = \frac{\sqrt{\hat{\sigma}_1}}{\sqrt{\hat{\sigma_i}^k + \hat{\sigma}_1}}
\end{equation}

To match the compact form in the main text, we adopt a notation simplification by redefining $\hat{\sigma_i}^k \rightarrow 2\hat{\sigma_i}^k$ (scaling the variance by 2, which preserves the mathematical nature of the Gaussian distribution). This leads to:
\begin{equation}
	\frac{\sqrt{\hat{\sigma}_1}}{\sqrt{2\hat{\sigma_i}^k + \hat{\sigma}_1}} \cdot \frac{1}{2} = \frac{\sqrt{\hat{\sigma}_1}}{2\sqrt{2\hat{\sigma_i}^k + \hat{\sigma}_1}}
\end{equation}

and the exponent term becomes:
\begin{equation}
	\exp\left( -\frac{(\hat{m}_i - \mu_i^k)^2}{2\hat{\sigma_i}^k + \hat{\sigma}_1} \right)
\end{equation}

\subsection{Final Result}
Combining all the above steps, we derive the fuzzy memory distribution as presented in Equation (9):
\begin{equation}
	\label{eq:p}
	Q(\hat{m}_i|y=k) = \frac{\sqrt{\hat{\sigma}_1}}{2\sqrt{2\hat{\sigma_i}^k + \hat{\sigma}_1}} \exp\left( -\frac{(\hat{m}_i - \mu_i^k)^2}{2\hat{\sigma_i}^k + \hat{\sigma}_1} \right)
\end{equation}

\begin{table*}[t]
	\centering
	\caption{Accuracy(\%) and adaptation time per instance (ms) on the Digits and Office31 datasets, where $\overline{Acc}$ and $\overline{Time}$ are the average accuracy and time cost across all tasks. $\dagger$ means the reproduced results.}
	\hspace{-0.42cm}
	\resizebox{1.03\textwidth}{!}{
	\begin{minipage}{0.39\linewidth}
		\centering
		
		\resizebox{\linewidth}{!}{
			\begin{tabular}{lccccc}
				\toprule
				Method & $S\rightarrow M$ & $U\rightarrow M$ & $M\rightarrow U$ & $\overline{Acc}$($\uparrow$) & $\overline{Time}$($\downarrow$) \\
				\midrule
				\rowcolor{gray!25}
				\multicolumn{6}{l}{\textbf{DAMap Methods}} \\
				w/o DA & 72.4 & 87.3 & 79.2 & 79.6 & - \\
				retrain@last & 77.1 & 92.1 & 91.6 & 86.9 & 1.213 \\
				retrain@BLS & 80.8 & 93.9 & 92.5 & 89.0 & 0.063 \\
				retrain@KNN & 77.6 & 91.7 & 92.1 & 87.1 & 0.044 \\
				retrain@DCT & 56.0 & 66.2 & 77.4 & 66.6 & 0.155 \\
				retrain@RF & 73.8 & 88.5 & 91.2 & 84.5 & 0.458 \\
				retrain@SVM & 75.5 & 91.6 & 92.6 & 86.6 & 3.720 \\
				retrain@BAG & 78.4 & 92.5 & 92.7 & 87.8 & 0.494 \\
				retrain@NBY & 81.1 & 89.5 & 91.8 & 87.5 & 0.022 \\
				retrain@XGB & 73.1 & 87.4 & 89.8 & 83.4 & 0.451 \\
				\textbf{MemFlow} & \textbf{82.5} & \textbf{93.8} & \textbf{91.2} & \textbf{89.1} & \textbf{0.013} \\
				\midrule
				\rowcolor{gray!25}
				\multicolumn{6}{l}{\textbf{SFUDA Methods}} \\
				SHOT & 98.9 & 97.5 & 98.0 & 98.1 & 0.091 \\
				AaD & 98.4 & 96.7 & 97.9 & 97.7 & 0.319 \\
				PFC & 98.5 & 97.8 & 98.0 & 98.1 & 1.056 \\
				TPDS & 98.9 & 98.0 & 98.4 & 98.4 & 3.362 \\
				\bottomrule
			\end{tabular}%
		}
		\label{tab:acc-digits}%
	\end{minipage}
	\begin{minipage}{0.57\linewidth}
		\centering
		\resizebox{\linewidth}{!}{
			\begin{tabular}{lcccccccc}
				\toprule
				Method & $A\rightarrow D$ & $A\rightarrow W$ & $D\rightarrow A$ & $D\rightarrow W$ & $W\rightarrow A$ & $W\rightarrow D$ & $\overline{Acc}$($\uparrow$) & $\overline{Time}$($\downarrow$) \\
				\midrule
				\rowcolor{gray!25}
				\multicolumn{9}{l}{\textbf{DAMap Methods}} \\
				w/o DA & 80.5 & 76.7 & 60.5 & 94.7 & 63.2 & 98.6 & 79.0 & - \\
				retrain@last & 84.3 & 79.8 & 60.6 & 94.2 & 64.7 & 99.0 & 80.4 & 1.397 \\
				retrain@BLS & 80.9 & 79.4 & 62.1 & 94.8 & 66.0 & 99.4 & 80.4 & 0.084 \\
				retrain@KNN & 90.0 & 84.8 & 61.9 & 95.0 & 62.8 & 99.6 & 82.3 & 0.046 \\
				retrain@DCT & 41.4 & 36.0 & 21.4 & 39.9 & 25.0 & 53.4 & 36.2 & 0.382 \\
				retrain@RF & 81.5 & 77.2 & 56.6 & 89.7 & 58.7 & 99.2 & 77.2 & 1.504 \\
				retrain@XGB & 70.9 & 64.4 & 39.6 & 72.0 & 41.4 & 88.6 & 61.4 & 0.636 \\
				retrain@SVM & 83.5 & 78.0 & 59.9 & 94.7 & 62.4 & 99.2 & 79.6 & 0.473 \\
				retrain@BAG & 87.3 & 87.0 & 62.3 & 95.6 & 64.1 & 99.4 & 82.6 & 0.753 \\
				retrain@NBY & 84.3 & 81.6 & 63.3 & 92.6 & 65.3 & 97.4 & 80.8 & 0.058 \\
				\textbf{MemFlow} & \textbf{92.2} & \textbf{88.9} & \textbf{68.2} & \textbf{97.5} & \textbf{70.4} & \textbf{99.4} & \textbf{86.1} & \textbf{0.160} \\
				\midrule
				\rowcolor{gray!25}
				\multicolumn{9}{l}{\textbf{SFUDA Methods}} \\
				SHOT & 94.0 & 90.1 & 74.7 & 98.4 & 74.9 & 99.9 & 88.7 & 4.682 \\
				AaD & 96.4 & 92.1 & 75.0 & 99.1 & 76.5 & 100.0 & 89.9 & 2.656 \\
				PFC & 97.3 & 94.0 & 75.6 & 99.2 & 76.6 & 100.0 & 90.5 & 3.901 \\
				TPDS & 97.1 & 94.5 & 75.7 & 98.7 & 75.5 & 99.8 & 90.2 & 25.230 \\
				\bottomrule
			\end{tabular}%
		}
		\label{tab:acc-office31}%
	\end{minipage}
}
	\label{tab:acc-digits-office31}
\end{table*}
\begin{table*}[t]
	\centering
	\caption{Accuracy(\%) and adaptation time per instance (ms) on the VisDA-C dataset, where $\overline{Acc}$ and $\overline{Time}$ are the average accuracy and time cost across all task. $\dagger$ means the reproduced results.}
	\resizebox{0.9\linewidth}{!}{
	\begin{tabular}{lccccccccrccccc}
		\toprule    Method & plane & bcycl & bus   & car   & horse & knife & mcycl & person & \multicolumn{1}{c}{plant} & sktbrd & train & truck & $\overline{Acc}$($\uparrow$) & $\overline{Time}$($\downarrow$) \\
		\midrule
		\rowcolor{gray!25}
		\multicolumn{15}{l}{\textbf{DAMap Methods}} \\
		w/o DA & 60.9  & 21.6  & 50.9  & 67.6  & 65.8  & 6.3   & 82.2  & 23.2  & \multicolumn{1}{c}{57.3 } & 30.6  & 84.6  & 8.0   & 46.6  & - \\
		retrain@last & 65.9  & 17.5  & 73.0  & 68.1  & 70.4  & 27.6  & 86.3  & 23.1  & \multicolumn{1}{c}{78.1 } & 45.9  & 79.8  & 14.1  & 54.1  & 2.375 \\
		retrain@BLS & 63.8  & 1.6   & 61.1  & 74.0  & 69.3  & 1.2   & 83.6  & 0.2   & \multicolumn{1}{c}{96.8 } & 10.9  & 80.2  & 1.9   & 45.4  & 0.096 \\
		retrain@KNN & 70.4  & 29.0  & 77.8  & 72.2  & 73.6  & 58.7  & 92.1  & 17.0  & \multicolumn{1}{c}{64.4 } & 14.6  & 80.6  & 5.8   & 54.7  & 0.045 \\
		retrain@DCT & 41.8  & 6.5   & 58.9  & 36.5  & 30.4  & 18.4  & 39.9  & 13.8  & \multicolumn{1}{c}{25.7 } & 19.4  & 64.1  & 20.2  & 31.3  & 0.21 \\
		retrain@RF & 60.0  & 13.8  & 76.7  & 58.7  & 44.4  & 29.7  & 77.8  & 16.5  & \multicolumn{1}{c}{40.1 } & 46.4  & 81.0  & 13.9  & 46.6  & 0.602 \\
		retrain@XGB & 52.0  & 10.1  & 73.3  & 56.9  & 45.2  & 23.1  & 70.6  & 19.5  & \multicolumn{1}{c}{39.9 } & 36.5  & 80.9  & 22.2  & 44.2  & 5.695 \\
		retrain@SVM & 63.4  & 17.1  & 73.0  & 63.9  & 62.1  & 53.1  & 85.3  & 20.2  & \multicolumn{1}{c}{34.9 } & 38.7  & 82.3  & 19.0  & 51.1  & 13.235 \\
		retrain@BAG & 66.9  & 19.4  & 73.9  & 78.1  & 75.8  & 60.0  & 93.1  & 15.1  & \multicolumn{1}{c}{66.7 } & 3.2   & 80.2  & 3.8   & 53.0  & 2.018 \\
		retrain@NBY & 64.5  & 58.0  & 81.6  & 51.7  & 68.3  & 17.0  & 52.5  & 57.6  & \multicolumn{1}{c}{64.6 } & 40.1  & 69.6  & 0.4   & 52.2  & 0.026 \\
		\textbf{MemFlow} & \textbf{68.4} & \textbf{56.0} & \textbf{75.0} & \textbf{67.1} & \textbf{80.3} & \textbf{69.3} & \textbf{83.5} & \textbf{44.3} & \multicolumn{1}{c}{\textbf{70.7}} & \textbf{12.9} & \textbf{80.3} & \textbf{16.9} & \textbf{60.4} & \textbf{0.012} \\
		\midrule
		\rowcolor{gray!25}
		\multicolumn{15}{l}{\textbf{SFUDA Methods}} \\
		SHOT  & 94.3  & 88.5  & 80.1  & 57.3  & 93.1  & 94.9  & 80.7  & 80.3  & \multicolumn{1}{c}{91.5} & 89.1  & 86.3  & 58.2  & 82.9  & 5.251 \\
		AaD   & 97.4  & 90.5  & 80.8  & 76.2  & 97.3  & 96.1  & 89.8  & 82.9  & \multicolumn{1}{c}{95.5} & 93    & 92    & 64.7  & 88.0  & 3.060  \\
		PFC   & 91.3  & 83.7  & 77.1  & 49.8  & 89.7  & 88.7  & 79.8  & 78.9  & \multicolumn{1}{c}{87.6} & 84.9  & 80.1  & 57.9  & 79.1  & 3.766 \\
		TPDS  & 97.6  & 91.5  & 89.7  & 83.4  & 97.5  & 96.3  & 92.2  & 82.4  &   96.0    & 94.1  & 90.9  & 40.4  & 87.6  & 67.288 \\
		\bottomrule    
		\end{tabular}%
	}
	\label{tab:acc-visda}%
\end{table*}%

\section{More Experimental Results}
\subsection{Detail Accuracy of Each Task}
\label{tab:full-acc}
To align with the evaluation protocol of existing SFUDA methods, the metric for the VisDA-C dataset was modified, which is shown in Table.\ref{tab:acc-visda}. The adaptation performance from the Synthetic to Real task $S\rightarrow R$ is evaluated using the mean per-class accuracy, while the reverse adaptation task $R\rightarrow S$ is no longer considered.

In addition, the detailed results about the accuracy of each task on Digits and Office-31 ares shown in Table\ref{tab:acc-digits} and Table.\ref{tab:acc-office31}, respectively. The experimental results demonstrate that MemFlow outperforms both traditional classifiers and the retrain-last-layers approach across all tasks, providing more detailed evidence of its distinct advantages in the DAMap scenario. Furthermore, compared with SFUDA methods, MemFlow achieves comparable performance while reducing time consumption by a significant margin. This remarkable performance effectively validates the feasibility of rapid domain adaptation on edge devices.

\begin{figure*}[t]
	\centering
	\hspace{-1cm}
	\resizebox{1.05\textwidth}{!}{%
		\begin{minipage}{\textwidth}
			\centering
			\begin{subfigure}[b]{0.23\linewidth}
				\centering
				\includegraphics[width=\linewidth]{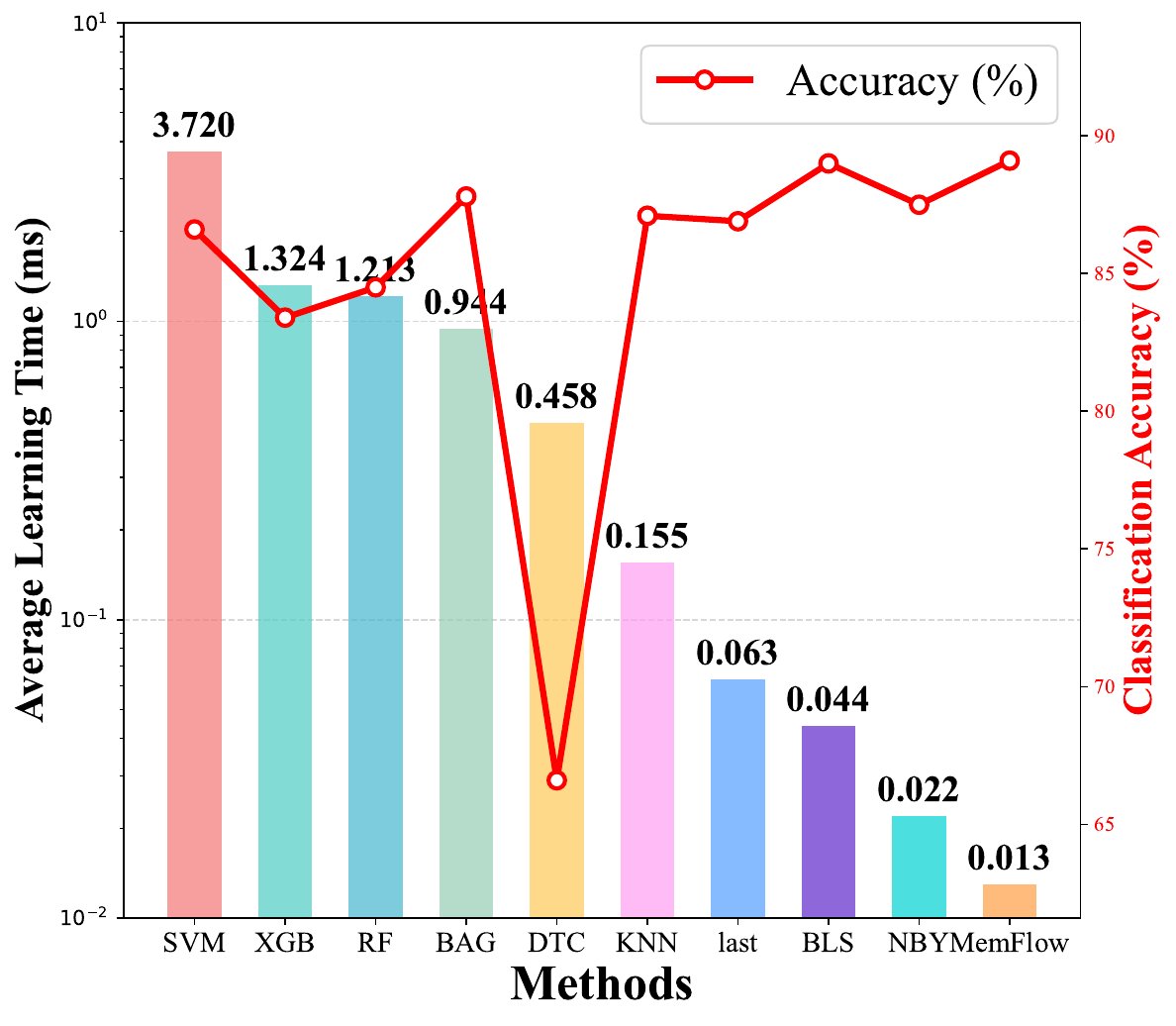}
				\caption{Digits}
				\label{acc-time-digits}
			\end{subfigure}
			\begin{subfigure}[b]{0.23\linewidth}
				\centering
				\includegraphics[width=\linewidth]{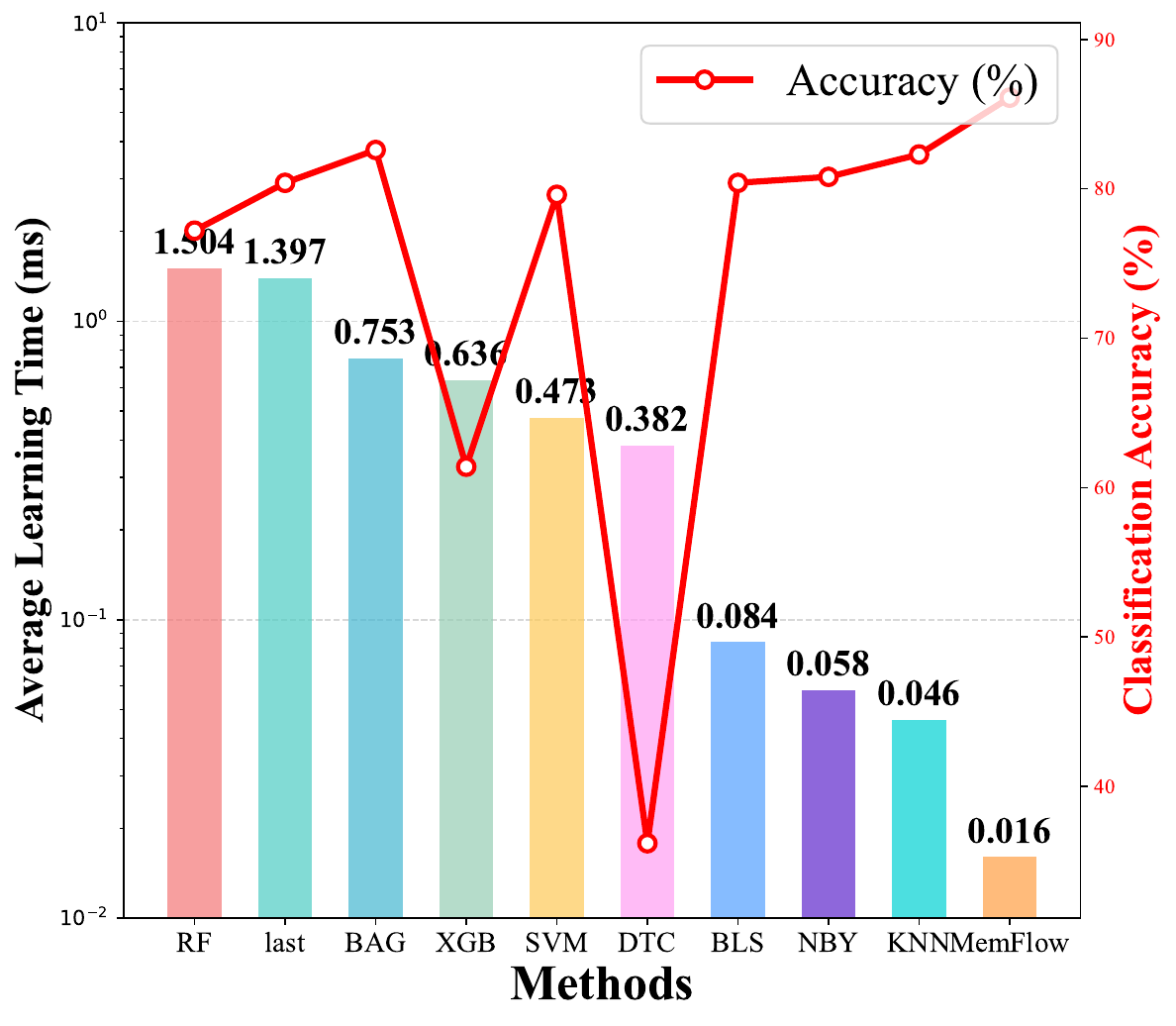}
				\caption{Office31}
				\label{acc-time-office31}
			\end{subfigure}
			\begin{subfigure}[b]{0.23\linewidth}
				\centering
				\includegraphics[width=\linewidth]{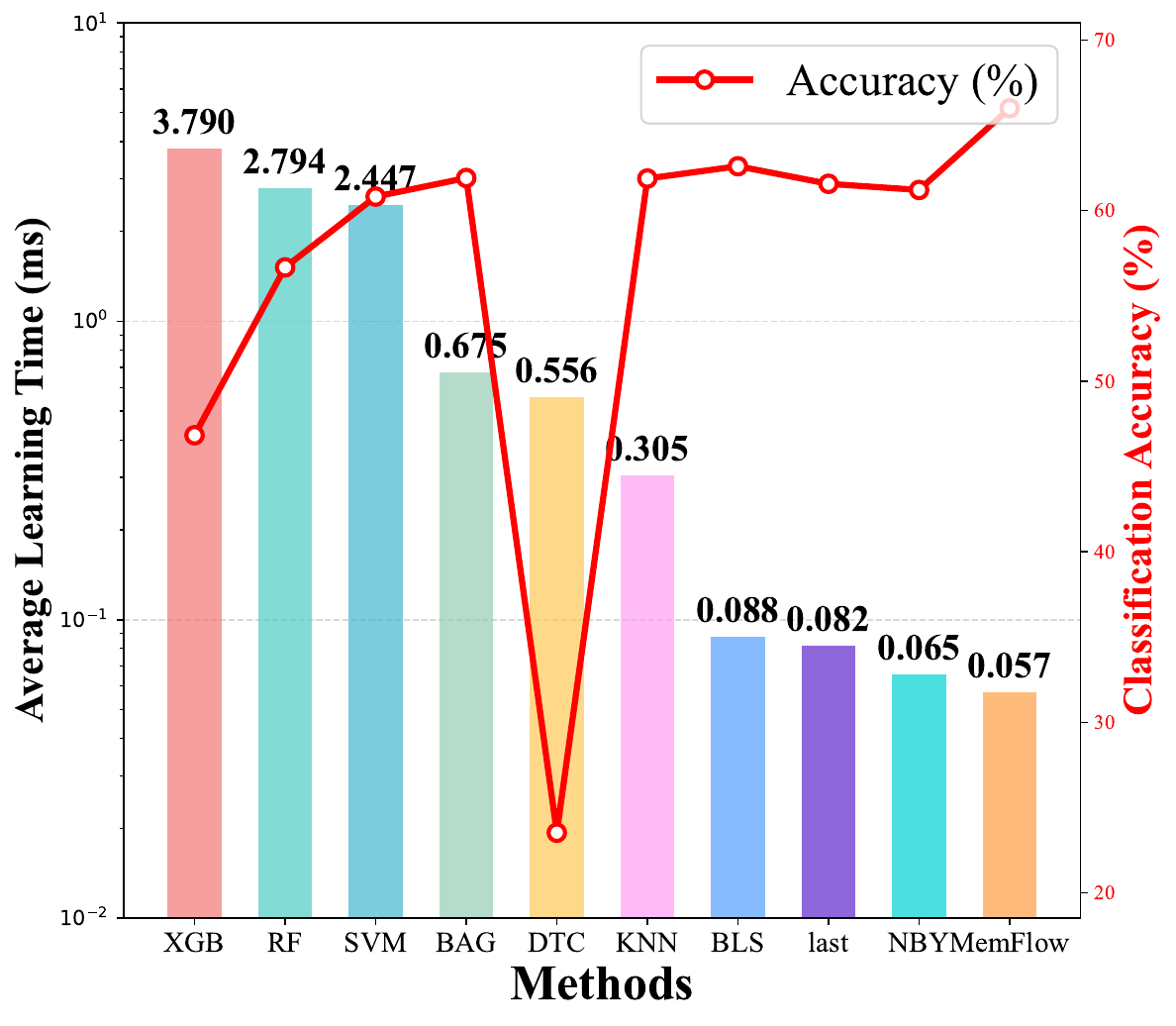}
				\caption{Office-Home}
				\label{acc-time-office-home}
			\end{subfigure}
			\begin{subfigure}[b]{0.23\linewidth}
				\centering
				\includegraphics[width=\linewidth]{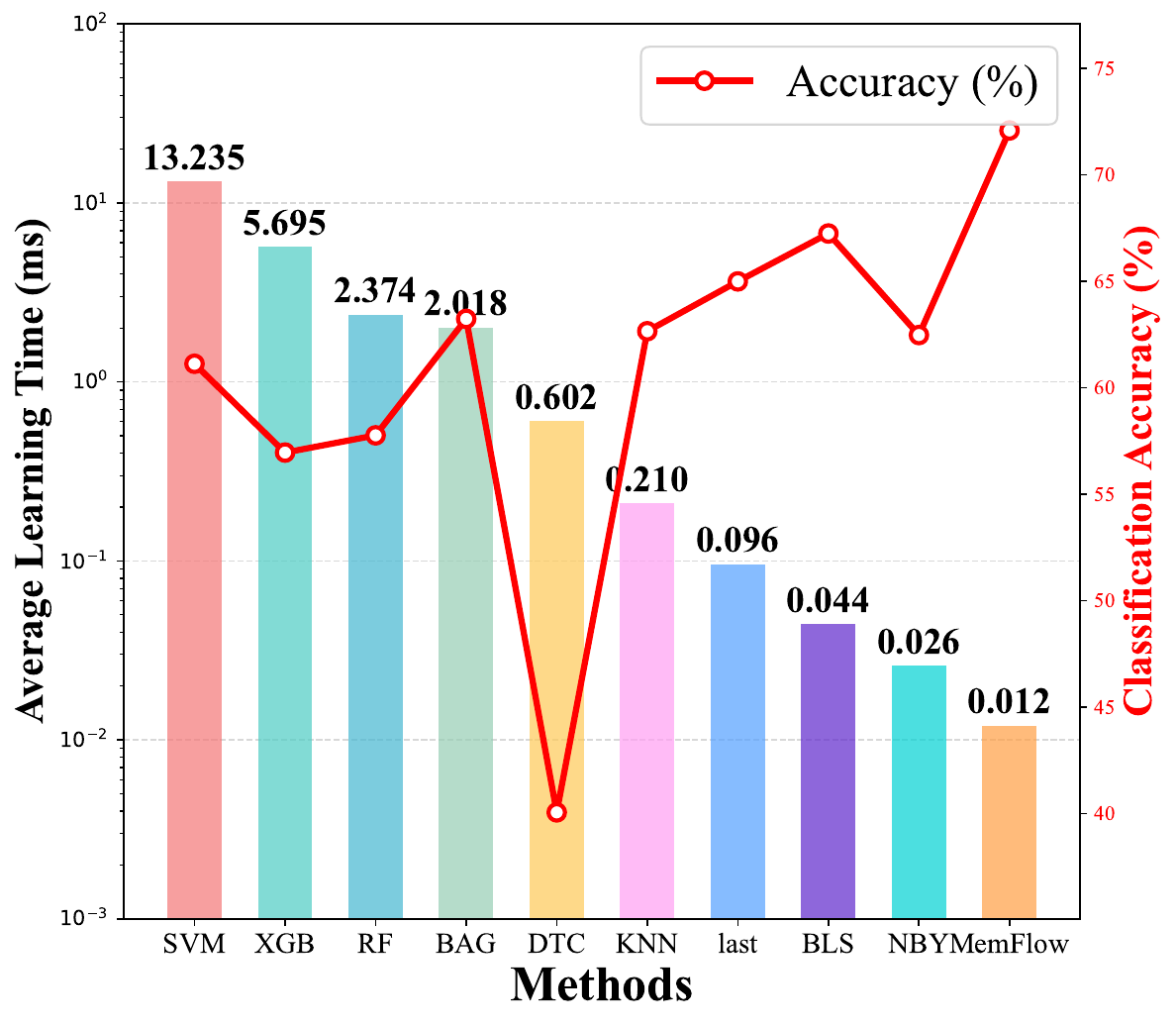}
				\caption{VisDA-C}
				\label{acc-time-visda}
			\end{subfigure}
		\end{minipage}%
	}
	\caption{Illustration of Accuracy and Time cost per instance across different methods on DAMap in all datasets.}
	\label{fig:acc-time-all-dataset}
\end{figure*}

To intuitively demonstrate the performance and efficiency advantages of MemFlow in the DAMap scenario, Figure \ref{fig:acc-time-all-dataset} summarizes the accuracy and time consumption across all datasets. The results show that MemFlow significantly surpasses all other classifiers. It achieves a substantial increase in accuracy while drastically reducing the time required compared to both the traditional classifier and the retrain-last-layers approach. Notably, MemFlow is even twice as fast as the lightweight retrain@NBY model, indicating its strong potential for real-time domain adaptation tasks.

%
\begin{figure*}[t]
	\centering
	\hspace{-1cm}
	\resizebox{1.05\textwidth}{!}{%
		\begin{minipage}{\textwidth}
			\centering
			\begin{subfigure}[b]{0.23\linewidth}
				\centering
				\includegraphics[width=\linewidth]{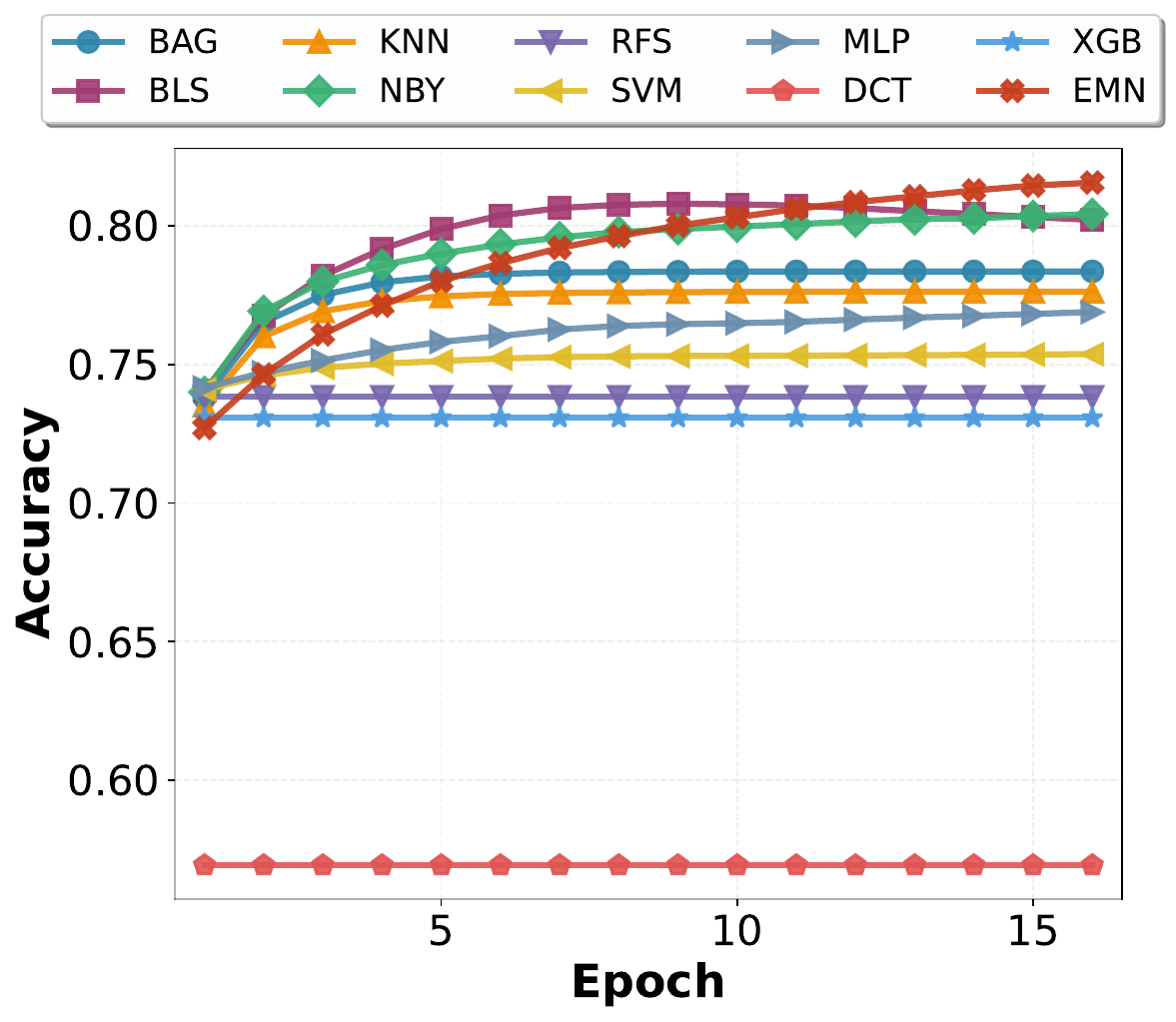}
				\caption{$S\rightarrow M$ in Digits Dataset}
				\label{acc-epoch-s2m}
			\end{subfigure}
			\begin{subfigure}[b]{0.23\linewidth}
				\centering
				\includegraphics[width=\linewidth]{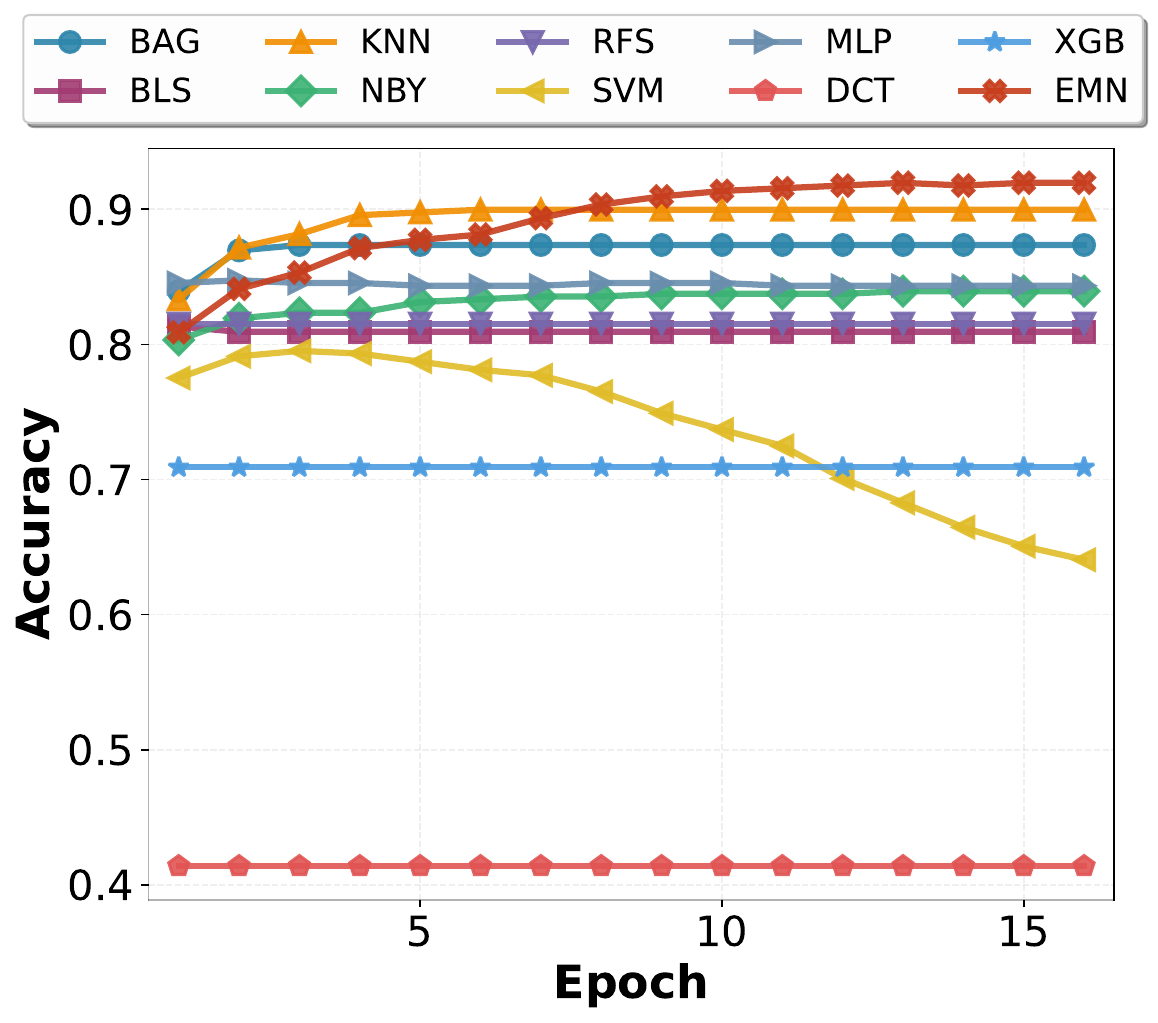}
				\caption{$A\rightarrow D$ in Office31 Dataset}
				\label{acc-epoch-a2d}
			\end{subfigure}
			\begin{subfigure}[b]{0.23\linewidth}
				\centering
				\includegraphics[width=\linewidth]{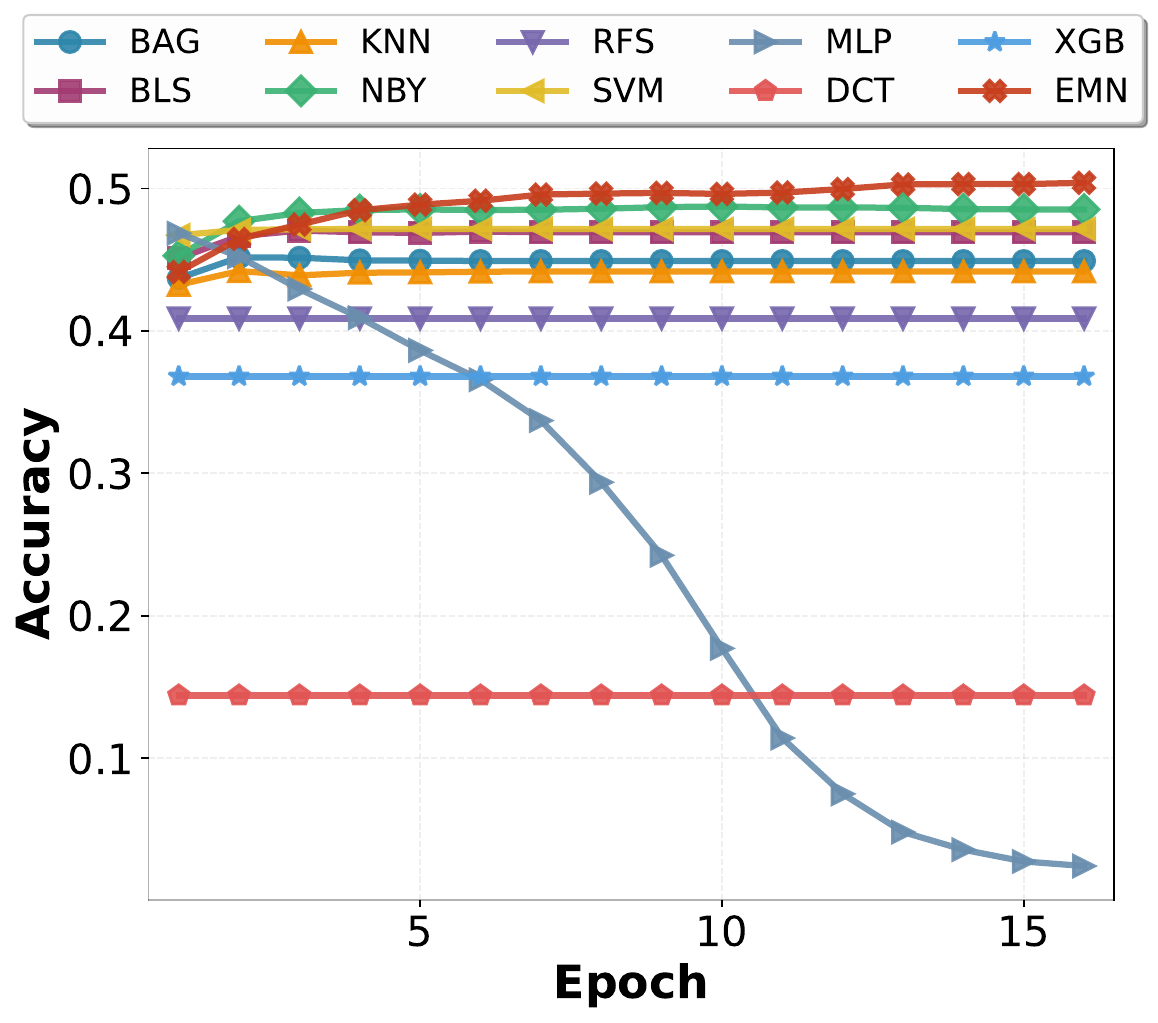}
				\caption{$A\rightarrow D$ in Office-Home Dataset}
				\label{acc-epoch-a2c}
			\end{subfigure}
			\begin{subfigure}[b]{0.23\linewidth}
				\centering
				\includegraphics[width=\linewidth]{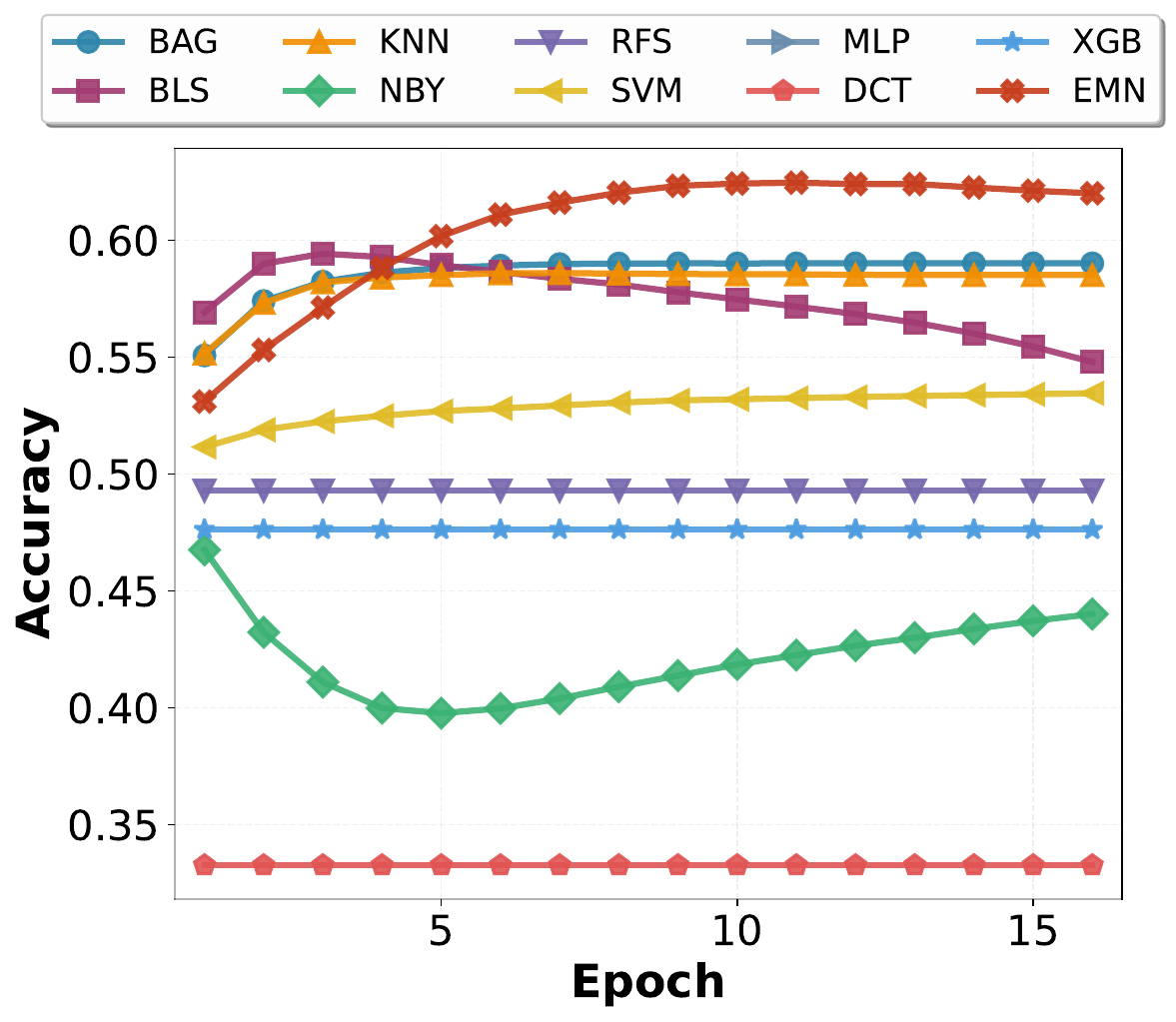}
				\caption{$S\rightarrow R$ in VisDA-C Dataset}
				\label{acc-epoch-s2r}
			\end{subfigure}
		\end{minipage}%
	}
	\caption{The accuracy trend on each epochs of domain adaptation in the different tasks with various datasets dataset.}
	\label{fig:acc-epoch}
\end{figure*}

\subsection{MemFlow in SFUDA Setting}
To demonstrate the effectiveness of MemFlow in the SFUDA scenario, we designed it as a plug-and-play module that can be integrated into existing SFUDA methods. Since MemFlow does not require gradient backpropagation, it serves as an instructor that assigns confidence scores to the pseudo-labels generated by the deep model. These confidence scores are then used to modulate the degree of gradient updates for each sample, thereby reducing the negative impact of incorrect pseudo-labels on the model.

Specifically, after the backbone network is pre-trained on the source domain, MemFlow rapidly memorizes the features extracted by the backbone, as detailed in Section 3.4. During training on the target domain, given the output features of deep neural model, it generates self-supervised prediction of samples as confidence distributions $\alpha_{Conf}$. Based on the confidence distribution $\alpha_{Conf}$, the downstream classification loss can be reformulated as the weighted loss: $L_{Conf} = \alpha_{Conf} * L.$

In this way, MemFlow leverages its ability to capture feature distribution shifts to assign reliability to pseudo-labels, effectively mitigating the influence of erroneous pseudo-labels.

Furthermore, MemFlow is updated only during the pseudo label generation process of the original SFUDA method. Through employing the Reinforced Memorization strategy described in Section 3.6, the mean $\mu$ and variance $\sigma$ of neurons in MemFlow are updated via momentum to adapt to the target domain distribution.

It is worth noting that since the AaD method does not originally include a pseudo-label generation step, integrating MemFlow requires adding such a process, leading to additional time overhead. In contrast, for other SOTA methods that already include pseudo-label generation, the inclusion of MemFlow introduces only a minimal time increase, approximately 1 ms.

\subsection{Accuracy in Different Epochs of DAMap}

Fig.~\ref{fig:acc-epoch} further depicts the accuracy of the models over the
epochs of learning pseudo labels on the target domain on each datasets. In all cases, MemFLow can improve the performance steadily, which benefits from the confidence-based parameter updating in Section 3.6. Notably, it can also be
observed that only retraining last layers (retrain@last) faces a significant drop in performance
in $A\rightarrow D$ of Office-Home Dataset, which may be caused by the accumulated errors of noisy pseudo labels. While the propose MemFLow can adaptively update the parameter to avoid the misleading of noisy pseudo labels.

\label{sec:epochs}

\begin{figure}[htbp]
	\vskip 0.2in
	\centering
	\includegraphics[width=1\linewidth]{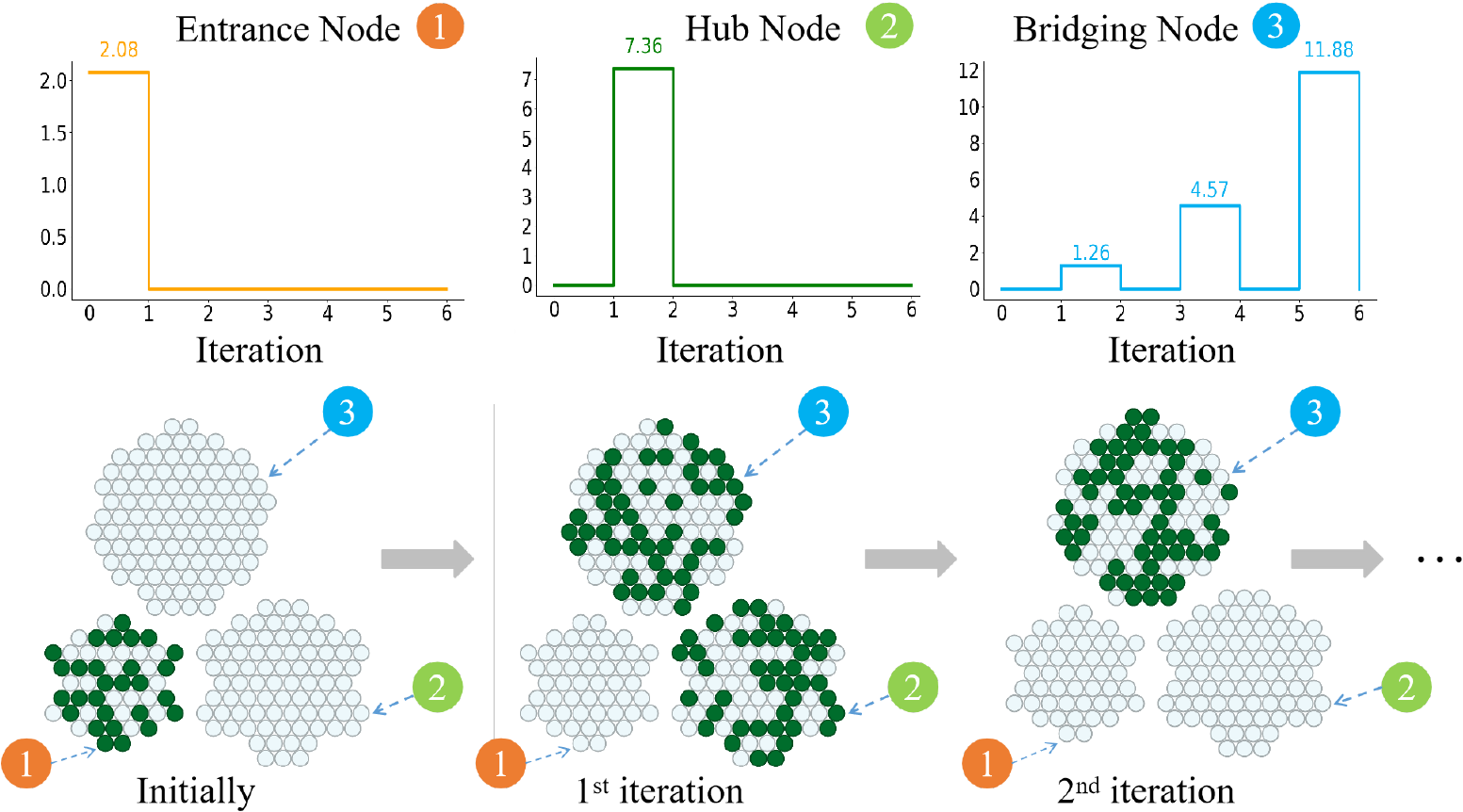}
	\caption{The change of the activation states and the output signals $o_{i,t}$ of the neurons in different iterations of signal propagation. For simplicity, we only show the nodes here while ignoring the connections. The green nodes indicate the activated ones with non-zero output.}
	\vskip 0.2in
	\label{fig:propagation}
\end{figure}
\begin{table*}[t]
	\centering
	\caption{Comparison with SNN Method in Office-Home Dataset, where $\overline{Acc}$ are the average accuracy across all task}
	\label{tab:SNN}
	\resizebox{\linewidth}{!}{
	\begin{tabular}{lccccccccccccc}
		\toprule
		 & $A\rightarrow C$
		& $A\rightarrow P$        & $A\rightarrow R$        & $C\rightarrow A$        & $C\rightarrow P$        & $C\rightarrow R$        & $P\rightarrow A$        &$P\rightarrow C$       & $P\rightarrow R$        & $R\rightarrow A$        & $R\rightarrow C$        & $R\rightarrow P$ & $\overline{Acc}$($\uparrow$) \\
		\midrule
		LIF\cite{knight1972dynamics} & 45.4 & 68.4 & 74.1 & 52.2 & 62.0 & 64.6 & 53.5 & 42.1 & 73.1 & 65.6 & 50.0 & 77.4 & 60.7 \\
		SRM\cite{gerstner1995time} & 40.4 & 63.6 & 68.0 & 44.7 & 58.2 & 60.2 & 44.8 & 37.9 & 66.9 & 60.2 & 45.2 & 75.0 & 55.4 \\
		ALIF\cite{brette2005adaptive} & 42.3 & 65.3 & 71.4 & 46.1 & 59.0 & 61.3 & 47.1 & 38.9 & 68.9 & 61.3 & 45.6 & 76.2 & 57.0 \\
		MemFlow+LIF & 46.1 & 63.2 & 67.4 & 48.1 & 59.1 & 63.8 & 50.4 & 47.3 & 76.5 & 65.8 & 51.7 & 77.0 & 59.7 \\
		MemFlow+SRM & 22.8 & 16.3 & 59.0 & 23.3 & 5.5 & 47.8 & 14.0 & 47.3 & 76.5 & 41.6 & 33.2 & 60.9 & 37.3 \\
		MemFlow+ALIF & 36.2 & 52.1 & 48.7 & 30.9 & 29.9 & 46.5 & 25.0 & 47.3 & 76.5 & 47.7 & 31.3 & 48.1 & 43.3 \\
		MemFlow & \textbf{50.4} & \textbf{76.5} & \textbf{76.9} & \textbf{59.3} & \textbf{71.1} & \textbf{69.7} & \textbf{59.9} & \textbf{47.3} & \textbf{76.5} & \textbf{69.5} & \textbf{53.9} & \textbf{81.0} & \textbf{66.0} \\
		\bottomrule
	\end{tabular}
}
\end{table*}

\subsection{Performance of Spiking Mechanism}
The  Spiking mechanism of MemFlow is illustrated in Fig. \ref{fig:propagation}. In contrast to traditional spiking neural networks, our spiking mechanism differs in that its output amplitude is dynamically tied to the magnitude of the positive hidden state (instead of fixed binary spikes), enabling fine-grained encoding of signal strength.
Furthermore, MemFlow integrates an explicit cumulative memory signal that transforms transient spiking activity into persistent traces, which yields a  long-term memory to support retrieval of input-related information. 

To further demonstrate the advantage of the proposed spiking procedure in MemFlow, We supply the comparison between the proposed model and other spiking neural networks, such as Leaky Integrate and Fire (LIF)\cite{knight1972dynamics}, Adaptive Leaky Integrate and Fire (ALIF)\cite{brette2005adaptive} and Spike Response Model (SRM) \cite{gerstner1995time}. Further, we replace the original spiking method in MemFlow with other spiking neural network. The experiments results are shown in Tab.\ref{tab:SNN}, which confirms the superior performance of the specific spiking design in MemFlow.

\begin{figure}[t]
	\centering
    \hspace{-0.5cm}
		\begin{minipage}{\linewidth}
			\centering
			\begin{subfigure}[b]{0.48\linewidth}
				\centering
				\includegraphics[width=\linewidth]{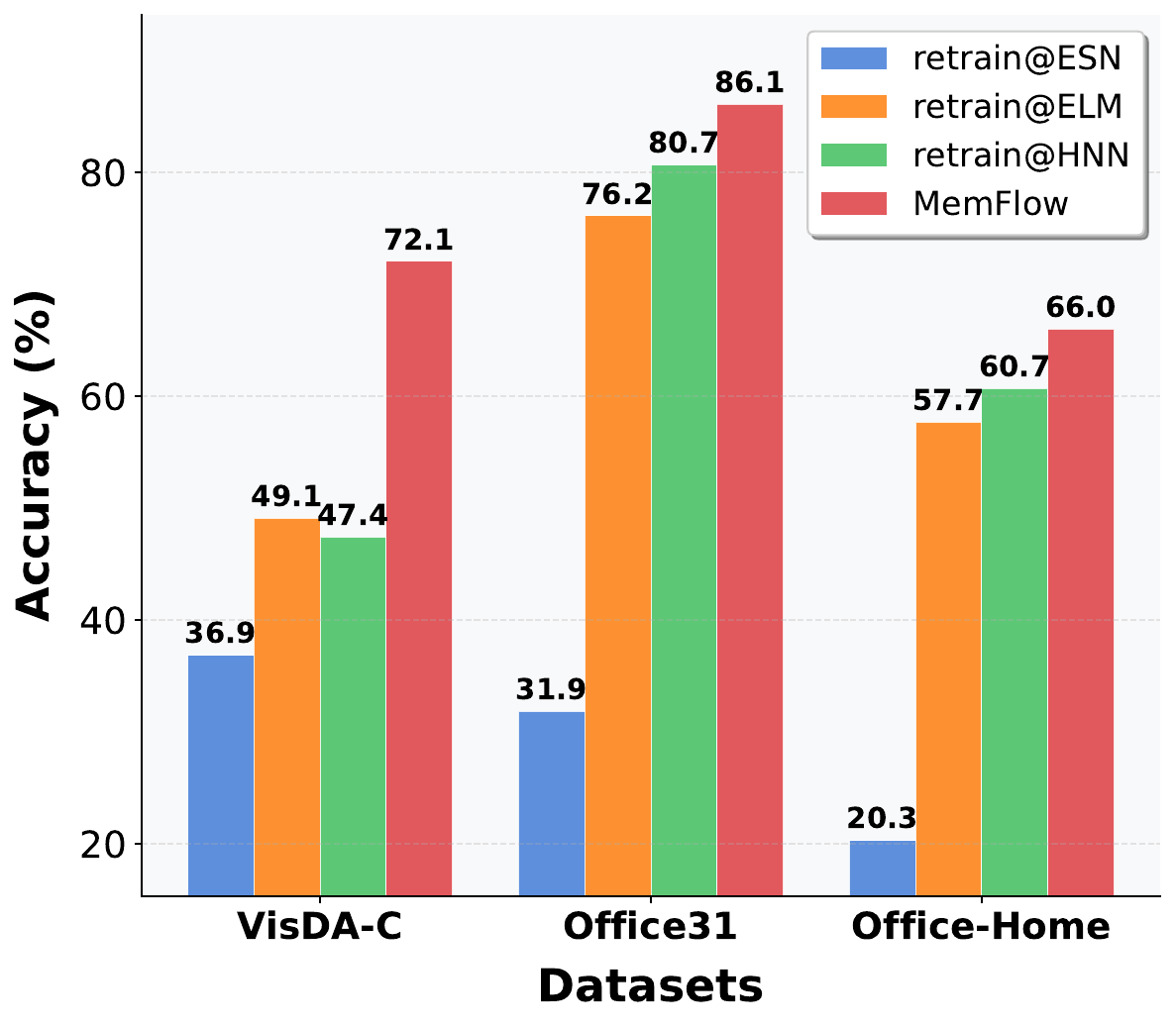}
				\caption{Accuracy}
				\label{ACC_No_Grad}
			\end{subfigure}
			\begin{subfigure}[b]{0.48\linewidth}
				\centering
				\includegraphics[width=\linewidth]{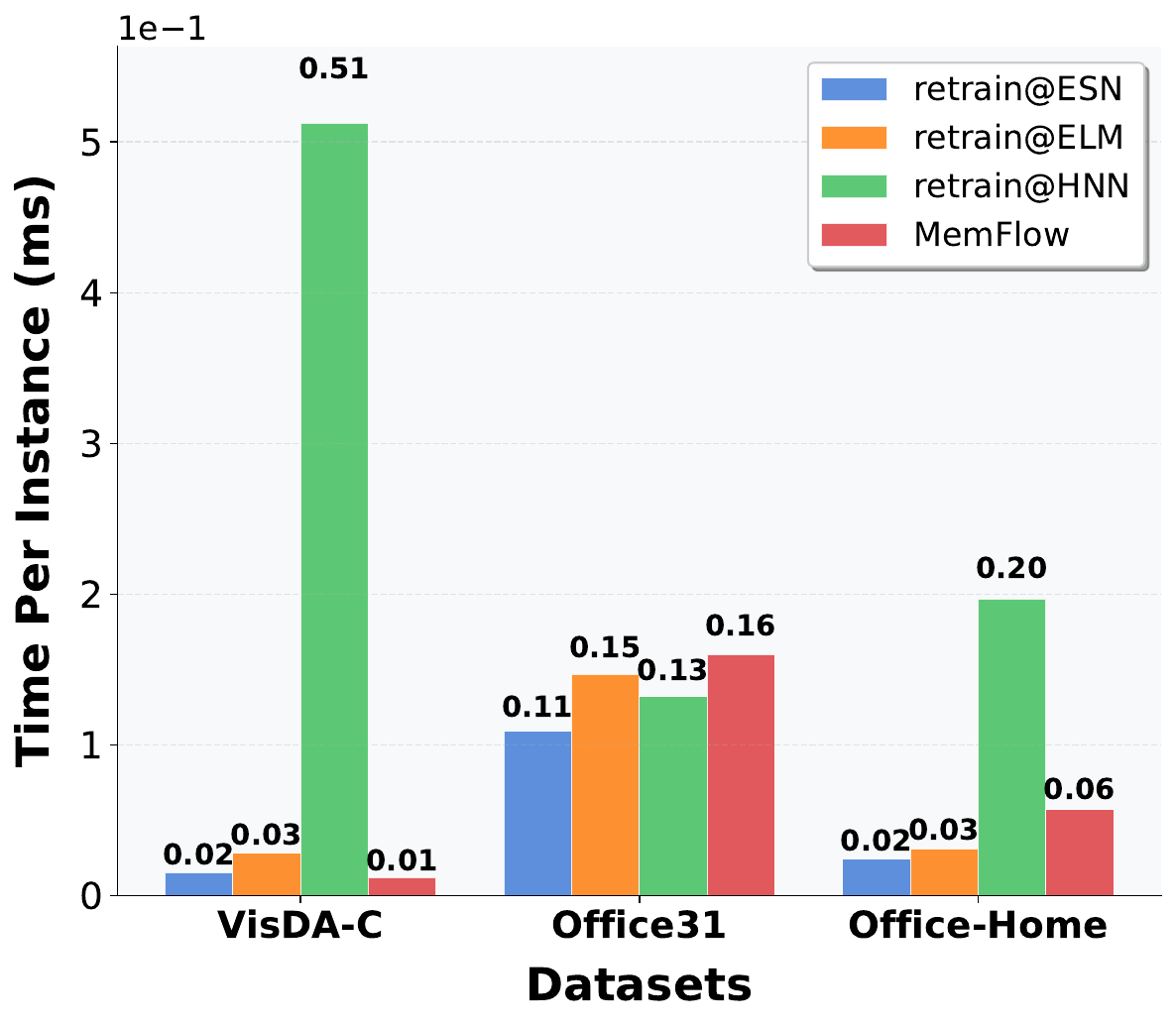}
				\caption{Time Per Instance}
				\label{Time_No_Grad}
			\end{subfigure}
		\end{minipage}%
	\caption{Comparison with Gradient-free Nerual Network in VisDA-C, Office31 and Office-Home Dataset.}
	\label{fig:No_Grad}
\end{figure}

\subsection{Comparison With Gradient-free nerual networks}
In this section, we compared the performance of the proposed MemFlow  against several Gradient-free nerual networks: 
\begin{itemize}
\item retrain@ELM is a single-hidden-layer feedforward Extreme Learning Machine\cite{huang2004extreme} network that uses random, fixed weights for the hidden layer and only trains the output weights analytically.
\item retrain@ESN\cite{jaeger2004harnessing} is a recurrent Echo State Network with a fixed, randomly connected reservoir of neurons whose dynamic state encodes the input history, and only a simple linear readout layer is trained.
\item retrain@HNN is a Hopfield Network\cite{hopfield1982neural}\cite{hopfield1985neural} that functions as an associative memory by converging to a stable state  closest to a given input pattern.
\end{itemize}
The experimental results, summarized in Fig.\ref{fig:No_Grad}, demonstrate the superior Performance of MemFlow. Crucially, while the computational time of MemFlow is comparable to that of these highly efficient models, it achieves a significant accuracy across all benchmark datasets. This key advantage stems from the fundamental difference in learning paradigm: whereas ELM, ESN, and Hopfield are fundamentally designed to fit a static input-output mapping function, MemFlow memorizes the associations within distributed neurons. This architecture enables a dynamic and reinforced memorization process, especially within the unlabeled target domain, allowing it to refine its internal representations and achieve greater generalization without a commensurate increase in computational overhead. 
\begin{figure}[h]
	\begin{center}       
		\includegraphics[width=1\linewidth]{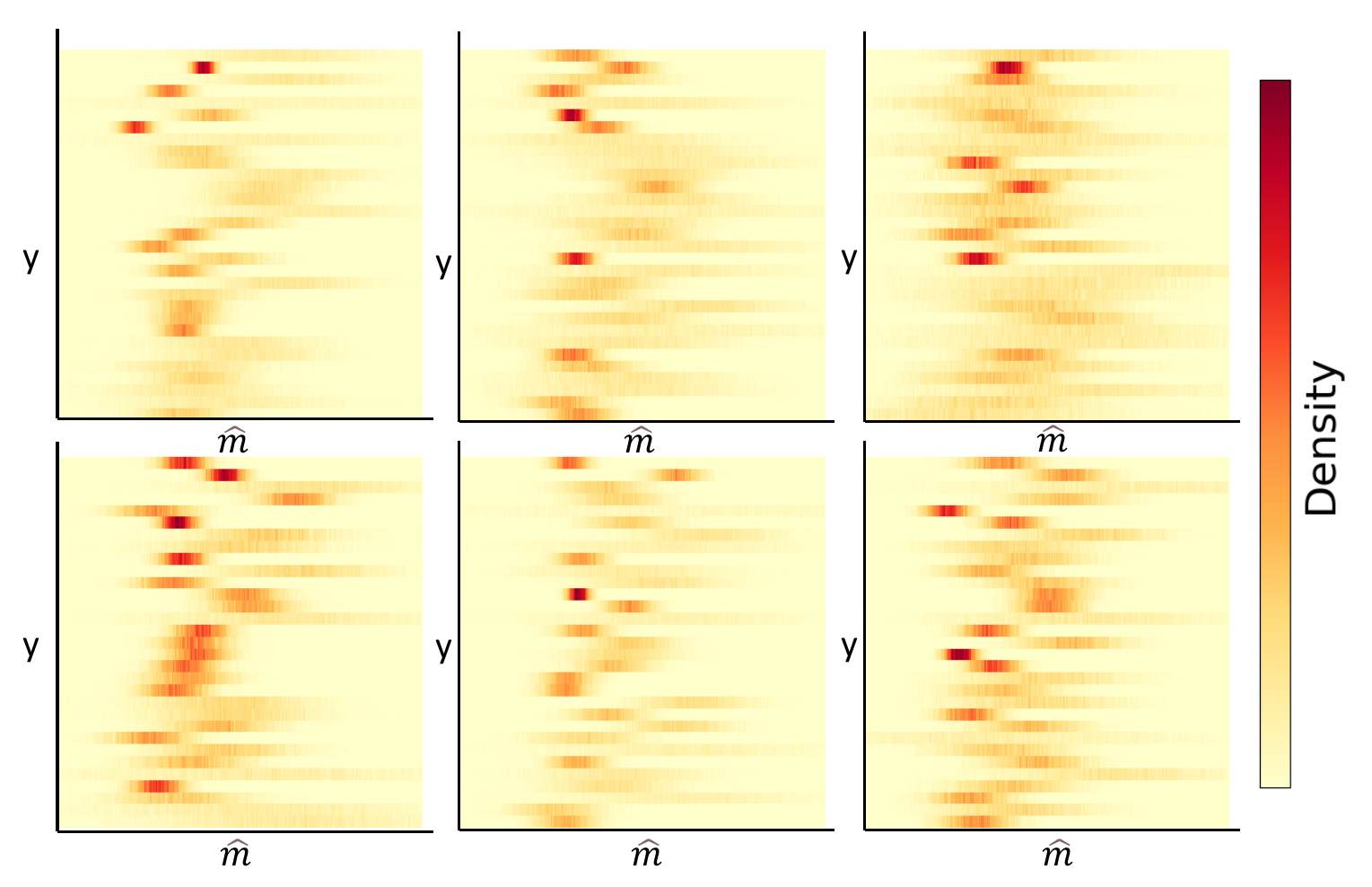}
	\end{center}
	\vspace{-0.3cm}
	
	\caption{Visualization of the memories on six randomly chosen neurons after training on the Office-31 dataset}
	\label{fig:memory_node}
\end{figure}

\subsection{Visualization}
Fig. \ref{fig:memory_node} depicts the memory units of randomly selected neurons, revealing various captured-distribution across different neurons in MemFlow. This diversity suggests that the different nodes exhibit specialized roles during the memory storage.


\end{document}